\documentclass[final,3p,times,twocolumn]{elsarticle}

\usepackage{amssymb}
\usepackage{graphicx}
\usepackage{caption}
\usepackage{amsmath,booktabs}
\usepackage{stmaryrd}
\usepackage[linesnumbered,ruled]{algorithm2e}
\usepackage{graphicx}
\usepackage{natbib}
\usepackage{textcomp}

\usepackage{multicol}
\usepackage{tikz}
\usepackage{lettrine}
\allowdisplaybreaks
\usepackage[export]{adjustbox}
\usepackage{ragged2e}
\usepackage{subfig}
\usepackage{float}
\usepackage{booktabs}
\usepackage{makecell, multirow}
\usepackage{tabularx}
\usepackage{siunitx}
\usepackage{threeparttable}
\newcolumntype{C}[1]{>{\centering\arraybackslash}m{#1}}
\usepackage{url}
\usepackage{notoccite}
\usepackage{placeins}

\biboptions{sort&compress}
\usepackage[colorlinks=true,linkcolor=black, citecolor=blue, urlcolor=blue]{hyperref}

\journal{Neurocomputing}

\begin{document}

\begin{frontmatter}


%% use the tnoteref command within \title for es;

\title{Ensemble plasticity and network adaptability in SNNs}

\author[label1]{Mahima Milinda Alwis Weerasinghe}
\ead{mahimarcrc@gmail.com}
\affiliation[label1]{organization={Department of Computer Science},
            addressline={Auckland University of Technology},
             city={Auckland},
             postcode={1010},
             country={New Zealand}}

\author[label3]{David Parry}
\ead{david.parry@murdoch.edu.au}
\affiliation[label3]{organization={Department of Computer Science},
	addressline={Murdoch University},
	city={Perth},
	postcode={6150},
	country={Australia}}

\author[label2]{Grace Wang}
\ead{grace.wang@aut.ac.nz}
\affiliation[label2]{organization={Department of Psychology and Neuroscience},
	addressline={Auckland University of Technology},
	city={Auckland},
	postcode={0627},
	country={New Zealand}}

\author[label1]{Jacqueline Whalley}
\ead{jacqueline.whalley@aut.ac.nz}

\begin{abstract}
Artificial Spiking Neural Networks (ASNNs) promise greater information processing efficiency because of discrete event-based (i.e., spike) computation. Several Machine Learning (ML) applications use biologically inspired plasticity mechanisms as unsupervised learning techniques to increase the robustness of ASNNs while preserving efficiency. Spike Time Dependent Plasticity (STDP) and Intrinsic Plasticity (IP) (i.e., dynamic spiking threshold adaptation) are two such mechanisms that have been combined to form an ensemble learning method. However, it is not clear how this ensemble learning should be regulated based on spiking activity. Moreover, previous studies have attempted threshold based synaptic pruning following STDP, to increase inference efficiency at the cost of performance in ASNNs. However, this type of structural adaptation, that employs individual weight mechanisms, does not consider spiking activity for pruning which is a better representation of input stimuli. In this study, we aimed to investigate spike regulation using ensemble learning in an information theoretic approach, to optimize ASNNs for spatiotemporal data classification. We envisaged that plasticity-based spike-regulation and spike-based pruning will result in ASSNs that perform better in low resource situations. In this paper, a novel ensemble learning method based on entropy and network activation is introduced, which is amalgamated with a spike-rate neuron pruning technique, operated exclusively using spiking activity. Two electroencephalography (EEG) datasets are used as the input for classification experiments with a three-layer feed forward ASNN trained using one-pass learning. During the learning process, we observed neurons assembling into a hierarchy of clusters based on spiking rate. It was discovered that pruning lower spike-rate neuron clusters resulted in increased generalization or a predictable decline in performance. Moreover, the networks demonstrated avalanche-like spiking activity under ensembled learning. The findings of this study draw attention to the ability of ensemble learning to push ASNNs towards criticality and of neuron pruning to allow ASNNs to ‘forget’ unimportant information. The results of this work illustrate how plasticity-based ensemble learning in conjunction with pruning can lead to increased robustness and efficiency of ASNNs in ML applications.              

\end{abstract}




\end{frontmatter}


\section{Introduction}
\label{Intro}
The emergence of technologies such as IoT (Internet of Things) have highlighted the need for data processing techniques that are efficient under low resource situations (i.e. limited power consumption and/or memory) \cite{Pfeiffer2018}. Quite often these techniques must deal with spatiotemporal data (STD) produced from a network of sensors. Therefore, such techniques need to consider preservation and use of vital information stored spatially and temporally during the processing. Artificial spiking neural networks (ASNNs) process data using discrete temporal events called spikes. According to the literature, these ASNNs promise greater information processing power and lower resource consumption than provided by traditional non-spiking artificial neural networks (ANNs) \cite{Maass1997a}. ASSNs have also been found to be ideal for spatiotemporal data processing since they are equipped with an inherent ability to code inputs using spikes on a temporal axis \cite{Pfeiffer2018,Tavanaei2019a,Roy2019}. In contrast, ANNs require specialized architectures, such as long short-term memory (LSTM) models, to process temporal data \cite{Greff2017}. These advantages of ASNNs have led to recent research efforts in the exploration and development of efficient ASNN algorithms for real-world Machine Learning (ML) applications involving STD \cite{Kasabov2013,Dora2016}. 

%Neurons and synapses are the fundamental building blocks of an Artificial Neural Network(ANN). Hence, both their function and, the quantity used has a direct impact on pattern recognition capability. In common ML approaches, neurons and synapses are identified separately, as processing and memory elements. Moreover, the network complexity is decided based on either: heuristics, intuition, search methods or motivations from literature. However, in biological systems, neurons are also found to be acting as memory substrates \cite{Desai1999,Zhang2003,Abraham2019}: contributing in the maintenance of network complexity through synaptic and non-synaptic plasticity \cite{Morris1999,Turney2012,Navlakha2018}. It is intriguing to question on how such functionality be adopted for ASNNs to leverage on the performance and efficiency found in biological systems such as the brain \cite{Javed2010}.

Biologically inspired ASSN learning strategies such as plasticity techniques allow for the local adaptation of synaptic and/or neuron properties using asynchronous spikes. These plasticity rules can be much more efficient than methods that depend on synchronous adaptation and, non-local transmission such as error backpropagation used in ANNs. Spike time dependent plasticity (STDP) \cite{Bi1998,Song2000,Abbott2000a} and intrinsic plasticity (IP) \cite{Desai1999,Zhang2003,Frick2005} are two such biologically plausible rules used in ASNNs for unsupervised learning. A handful of studies have used these plasticity rules ensembled in ML applications for static image data classification and, reported performance at par with ANNs \cite{Diehl2015b,Hao2020}. However, it is unclear how this ensemble learning should be regulated based on spiking activity to achieve robust performance with STD. Spike-based ASNN regulation is important in enabling efficient learning and, can be vital in online and incremental learning applications where spike-regulation based on final inference can be too time consuming.  

Neuronal pruning has been identified as a key process that eliminates extra or unnecessary synapses and/or neurons in the biological brain that increase the network efficiency \cite{Navlakha2018}. Inspired by the biological findings, ASNNs have adapted pruning in resource restricted applications \cite{Shi2019}. Past studies of neuronal pruning have been following \emph{proliferate then prune} approach for individual synapses \cite{Rathi2019,Shi2019}. However, comparing individual synaptic weights to a threshold may cause scalability issues. A probable solution to this would be neuron pruning based on spiking characteristics. Although bio-physical experiments have found such neuron pruning (i.e, apotheosis) to be helpful in network efficiency and pattern recognition \cite{Iglesias2006}, less is known about the effects of such mechanisms on ML applications with ASNNs. 

As a result of exploring the literature on ASNNs (reported in section 2) two gaps in knowledge were identified: 
\begin{enumerate}
\item The insufficiency of theoretically justified methods for spike regulation in ensemble learning  
\item The lack of understanding in terms of neuron pruning to increase efficiency  
\end{enumerate}

These gaps led to the formation of two central hypotheses in terms of ensemble learning and network adaptability. For ensemble learning, if all the neurons are activated and entropy was minimized (i.e., measure of uncertainty), then the ASNN would perform better in terms of robustness and efficiency; conjectured based initial work of Lazar \cite{Lazar2007} and, findings of our previous study \cite{Weerasinghe2021} that discussed the importance of increasing neuron activation for enhance pattern recognition. Secondly, if neurons of less spiking rate are pruned then, the performance of the ASNN trained with ensemble learning will gradually reduce ; conjectured based on the work of Iglesias and Savin \cite{Iglesias2006,Savin2010} that demonstrated hierarchical formation of neuron clusters based on input features. To test the hypotheses, we employed a feed-forward ASNN architecture, with a hidden layer trained using a single epoch. The performance was tested in a rigorous ML framework to understand training and inferring performance separately. The algorithm is used to classify Electroencephalography (EEG) data, a form of STD. In this study we make the following contributions:

\begin{itemize}
\item The introduction of a novel ensemble learning method based on information theory.  

\item The presentation of a novel neuron pruning method based on spiking rate. 

\item An exploration of avalanche-like spiking dynamics that may contribute to better pattern separation. 
\end{itemize}

The rest of this article is organized as follows: Section 2 presents related work, Section 3 introduces the methodology and experimental framework, Section 4 elucidates the results, Section 5 discusses the results and relates them to previous work and, Section 6 concludes with limitations \& future work. It should be noted that for the rest of the article the terms plasticity and learning are used interchangeably.

\section{Related work}
\label{Section 2}

\subsubsection{Standalone IP}
\label{IP Neuron models}
IP was initially investigated in vitro by Stemmler and Koch \cite{Stemmler1999} and, presented an unsupervised IP rule using Hodgkin-Huxley(HH) neuron \cite{Hodgkin1952}. This application of IP involved adjusting voltage-dependent conductance via the properties of voltage-gated ion channels. The authors elucidated the impact of IP on output firing probability distribution and its utility in maximizing mutual information. Since exponential distributions yields the highest entropy from all distributions for a non-negative random variable (i.e. given the mean is stationery), Triesch introduced an IP rule to reduce Kullback-Leibler(KL) divergence between desired exponential and actual firing probability distributions \cite{Triesch2007}. In a similar approach, Weibull distribution was considered instead of exponential \cite{Li2011}. Both these studies were based on non-spiking neurons and computer generated data. Other researchers have extended the idea of information maximization and studied the effect of IP on SNNs  \cite{Li2013,Li2016,Zhang2019,Zhang2019a}. Real-world applications of image recognition and speech classification was presented by Zhang and Li  \cite{Zhang2019} in which IP was used with a Liquid State Machine (LSM). The synaptic weights of the LSM were fixed and, IP was tuned using $R$ resistance and $\tau_{m}$ membrane time constant of each neuron to achieve a desired exponential firing distribution following the KL divergence method \cite{Triesch2007}. This method was modified to adjust intrinsic properties of the neurons according to backpropagated error gradient \cite{Zhang2021a}. In these studies, the weights are set randomly and kept constant \cite{Zhang2019} or changed according to backpropagation \cite{Zhang2021a} where learning rules are only dependent on the spiking rate. 

\subsubsection{IP and STDP - Ensemble Learning}
\label{IP with STDP}
STDP is an unsupervised learning mechanism used for synaptic weight update based on spike timing. Effects of applying STDP with IP was discussed by Savin, Joshi and Triesch \cite{Savin2010} . The authors reported neurons adapting to be responsive to independent components from the input signal (i.e. ICA-like learning) and, presented successful de-mixing of Gaussian signals. Another implementation of IP in a Liquid state machine (LSM) setup used Inter-Spike-Interval(ISI) for neuronal threshold tuning and, reported better performance with ensemble method over standalone STDP \cite{Li2018}. A simplistic model of IP is presented in combination with STDP by Lazar \cite{Lazar2007},which reported the advantages of such methods in time series prediction and highlights the importance of neurons operating in criticality for better performance.  In this study, IP was implemented using the equation,

\begin{equation}
	\label{EqLazar}
	T_{i}(t + 1) = T_{i}(t) + \eta_{IP}(x_{i}(t) - k/N)
\end{equation}

This IP algorithm increases threshold if a spike has occurred in the previous time step $T_{i}$ and decreases if not. This forces every unit in the network to be sensitive to the incoming stimuli. $\eta_{IP}$ is the small IP learning rate, $x_{i}(t)$ denotes the state (i.e. 1 or 0) at time step $t$. The total number of neurons in the network is $N$ and $k$ is decided based the desired number of neurons should be active at a given time step.
In an ML setup, STDP and IP ensemble learning was discussed in \cite{Diehl2015b,Hao2020}, where  authors reported state of the art accuracy levels. However, these studies have not been implemented to discuss the effects of ensemble learning at a spike level nor to explore the impact of such methods at different stages of training and testing. This led us to the question as to how the ensemble learning should be regulated at a spike level for a better performance in ML.  
 
\subsection{Structural adaptation in ASNNs}
In this section, we define any form of network architecture change as a structural adaptation technique. This includes adding and/or pruning neurons and/or synapses. Biologically, the brain is believed to go through a pruning after proliferation strategy \cite{Morris1999}, and this process has been observed in-vivo \cite{Turney2012}. Presumably, it is much likely for a brain circuit to get refined for better efficiency and robustness over time. However, most SNN applications in ML can be categorized as methods implemented either to enhance robustness\cite{Wysoski2008,Wang2014,Dora2016,Roy2017} or efficiency\cite{Rathi2019,Shi2019} separately. 

One learning algorithm based on neuronal growth was introduced specifically for performing image recognition \cite{Wysoski2008}. During the training period, at each passing of a sample, a new neuron map is evolved at the final layer. Researchers have not implemented a temporal learning method such as STDP. A similar approach involved adding neurons to the hidden layer in a winner-takes-all circuit to make a selected neuron fire first for a given class of sample during training \cite{Wang2014}. STDP and anti-STDP rules were used between output and hidden layers of the network. In another study, a two layer SNN with a rank order rule (RO) for weight updating was used to add a new neuron at the output layer for each training sample \cite{Dora2016}. Synaptic rewiring under STDP was introduced in separate study where a connection matrix is updated based on dynamic threshold \cite{Roy2017}. Synaptic rewiring takes place after each training epoch. These methods have not been implemented to deal in multi-spike environment or with a consideration on biological plausibility. This may affect the scalability and adaptability  of the network \cite{Navlakha2018} and, restrict the algorithm from making use of different neuronal dynamics for efficient coding. Therefore we intended to follow a pruning after proliferation strategy which is biologically plausible.    

Pruning based techniques of structural adaptation in the literature focuses on elimination of synapses to increase processing and implementation efficiencies. Rathi, Panda and Roy presented threshold based synaptic pruning where weights are updated according to STDP rule \cite{Rathi2019}. The threshold is a fixed value determined after evaluating the accuracy of the SNN. The same study presents the effects of weight quantization which adjusts all weights to a fixed range of values after training is completed. A different form of structural adaptation was presented by Shi, where synapses are assigned to their lowest value based on the efficacy decided by a threshold at each epoch \cite{Shi2019}. This leaves the possibility for a neuron to get activated at a later epoch, and synaptic elimination takes place only at the end of a training session.

The dynamic spiking activity produced by a neuron represents the input stimuli. This representation is made possible by synaptic (i.e, STDP) and non- synaptic plasticity (i.e, IP) in an ensemble learning setup. Therefore, we propose the spiking activity of a neuron to be a better representation of the input stimuli rather than the magnitude of individual synaptic weight. Hence, in contrast to the methods discussed, we focused on pruning neurons instead of individual synapses which is arguably more efficient in terms of computational cost. This form of programmed neuronal cell death known as apoptosis is observed in biology \cite{Iglesias2006}.

\section{Methodology and Data}
\label{Section 3}

\begin{figure}[t]
	\includegraphics[width=0.5\textwidth]{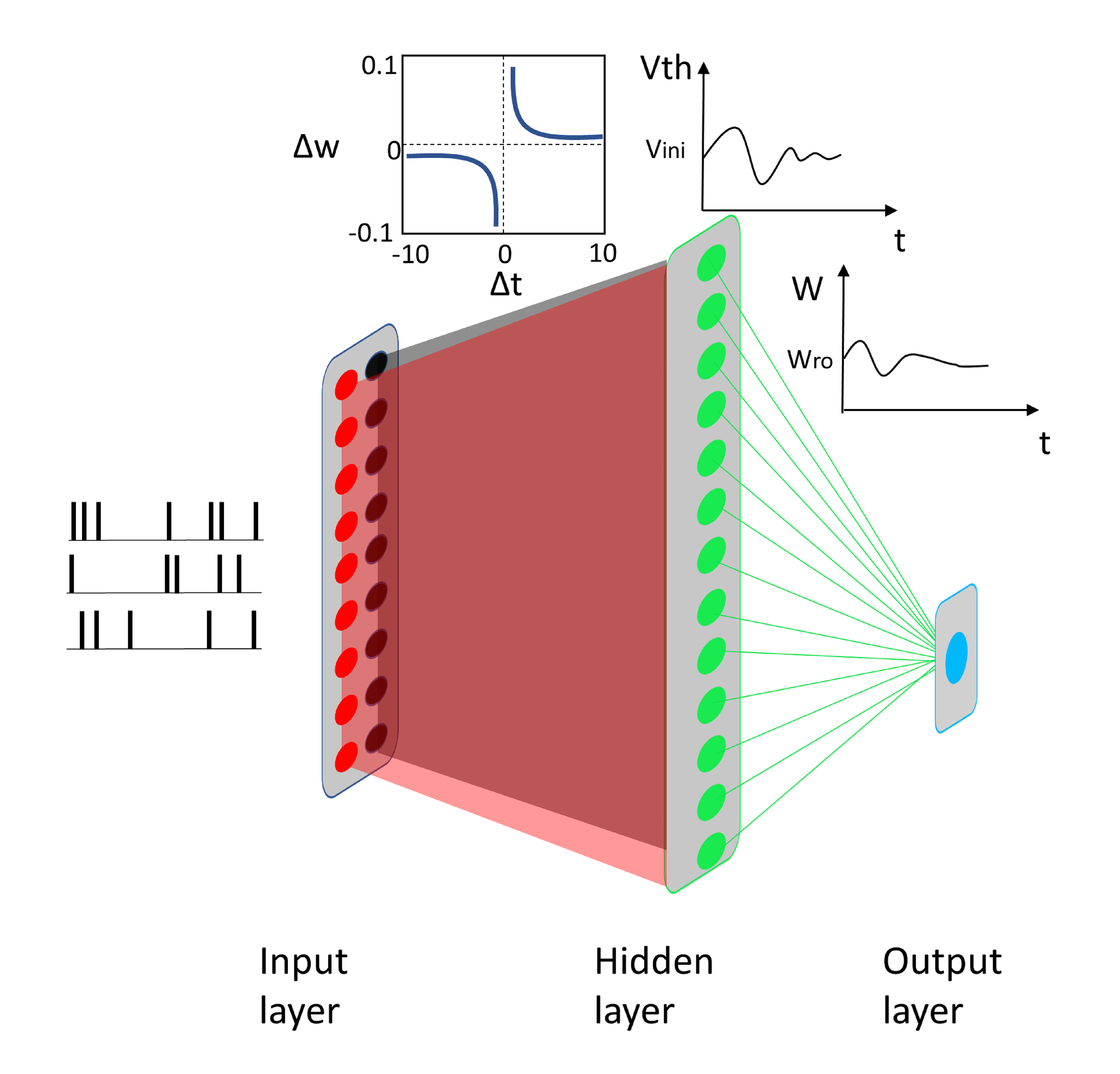}
	\captionof{figure}{SNN algorithm used for experimentation with a single hidden layer. The input layer consists of LIF neuron pairs capable of propagating both excitatory and inhibitory spikes (Red-excitatory and black-inhibitory). Synapses between input and hidden layer are connected and updated using STDP \cite{Song2000}. The hidden layer neurons undergo a threshold adaptation during the training. Based on the spiking rate during the training cycle, the hidden layer is pruned. The output layer evolves each time a sample is trained or tested. Hidden-to-output synapses are all excitatory where weight initiation is based on RO.}
	\label{TheNetwork}
\end{figure}

In order to investigate ensemble learning and neuron pruning, we employed a feed-forward SNN architecture(see Figure \ref{TheNetwork}). This network is equipped with threshold adaptable LIF neurons, a combination of STDP and IP for unsupervised learning, and a classifier based on semi-supervised learning. The network is trained under batch mode with one pass learning. 

\subsection{Leaky Integrate and Fire Neuron} 
The hidden layer of the SNN consists of LIF neurons, the behavior of which can be mathematically modeled using a resistor-capacitor circuit \cite{Gerstner2002a} (see page 95). We selected this model due to its processing efficiency and simplicity over more biologically plausible counter parts such as HH model \cite{Izhikevich2004}. The model introduced by Izhikevich \cite{Izhikevich2003a} is also capable of demonstrating many types of spiking behavior with low computational cost. However, this model consists of four independent parameters and two differential equations. Since our main objective was to investigate ensemble learning and network adaptability, we opted for the more simplistic, computationally efficient solution which is the LIF. Moreover,unlike a regular LIF use, since our implementation involves IP, we expected more dynamism in firing types \cite{Izhikevich2004}. 

\begin{equation}
	\label{EqLIF}
	\tau_{m} \frac{dv_{t}}{dt} =  v_{rest} - v_{t} + RI_{t} 
	\text{    where }\tau_{m} = R.C
\end{equation}

Here, the time constant $\tau_{m}$ is the multiplication of resistance $R$ and the capacitance $C$. The membrane potential $v_{t}$ change with the time dependent input current $I_{t}$, given that the neuron is not within the refractory period. During refractory the voltage resets to $v_{rest}$. Moreover, the spike generation of neuron is given in the form of a dirac delta function. 

\begin{equation}
	\label{EqDdelta}
	\delta(s_{t})=\begin{cases}
		1, \&  \text{ if }v_{t}=v_{thresh}.\\
		0, \&  \text{ otherwise}.
		\end{cases}
\end{equation}

A firing $s_{t}$ in the neuron takes place if the membrane potential $v_{t}$ reaches threshold voltage $v_{thresh}$. A firing is represented by 1 and non-firing by 0.
  
\subsection{Spike Time Dependent Plasticity}
We used the standard form of STDP \cite{Bi1998,Song2000,Abbott2000a} to update weights between the input and the hidden layer. This  unsupervised method of learning can be used effectively to learn from local spiking patterns \cite{Taherkhani2020}. In literature, we find other forms of STDP such as anti-STDP \cite{Wang2014}, power law weight-dependent STDP \cite{Rathi2019} and Dopamine modulated-STDP \cite{Hao2020}. We decided to proceed with the basic form of STDP which is the foundation of all the other methods, enabling the findings to be more generalizable. One of the key implications of STDP  plasticity is it's tendency to produce ``runaway synaptic potentiation'' where synapses get caught up in potentiation or depreciation loop which destabilizes the network \cite{Chen2013}. The spikes produced under such situations are not truly representative of the input stimuli. Previous literature suggests IP to be a remedial action against such destabilization \cite{Diehl2015b}.

\begin{equation}
	F({\Delta t})=  A_{+} \exp^{(-\Delta t/ \tau_{pos})} \Delta t > 0
	\label{EqApos}
\end{equation}
\begin{equation}
	F({\Delta t}) = -A_{-} \exp^{(\Delta t/ \tau_{neg})}  \Delta t < 0
	\label{EqAneg}
\end{equation}
\begin{equation}
	\Delta W_{ij} = \sum_{a}^{b}\sum_{p}^{q} F(t_{i}^{m} - t_{j}^{n})
	\label{EqSynW}
\end{equation}

In \ref{EqApos} and \ref{EqAneg}, the function $F({\Delta t})$ refers to long term potentiation (LTP) and depreciation (LTD) respectively. LTP takes place if the post-synaptic neuron fires after the pre-synaptic neuron where time gap between the firings, $\Delta t$ becomes positive. It becomes negative if the firing sequence happens vise-versa, leading to LTD. $\tau_{pos}$ and $\tau_{neg}$ denote the positive and negative time windows respectively for the synaptic modification. Factors $A_{+}$ and $A_{-}$ are used for positive and negative synaptic modification. The cumulative weight change is denoted by $W_{ij}$ where post-synaptic firing is considered from $a$ to $b$ and pre-synaptic from $p$ to $q$. Firing at a given time $m$ by post-synaptic neuron is given as $t_{i}^{m}$ and pre-synaptic neuron firing at time $n$ by $t_{j}^{n}$.  
 
\subsection{Intrinsic Plasticity}
\label{IP}
Inspired by the IP implementation introduced by Lazar \cite{Lazar2007}, we took a similar approach to implement a IP learning mechanism to operate with STDP. However, instead of the single learning rate introduced in \ref{EqLazar}, we used two separate learning rates  $\theta_{pos}$ and $\theta_{neg}$. This enabled us to increase or decrease $v_{thresh}(t)$ at different rates based on incoming spiking activity providing more flexibility around threshold adjustment.  
  
\begin{figure}
	\begin{center}
	\subfloat[]{%
		\includegraphics[width=0.5\linewidth]{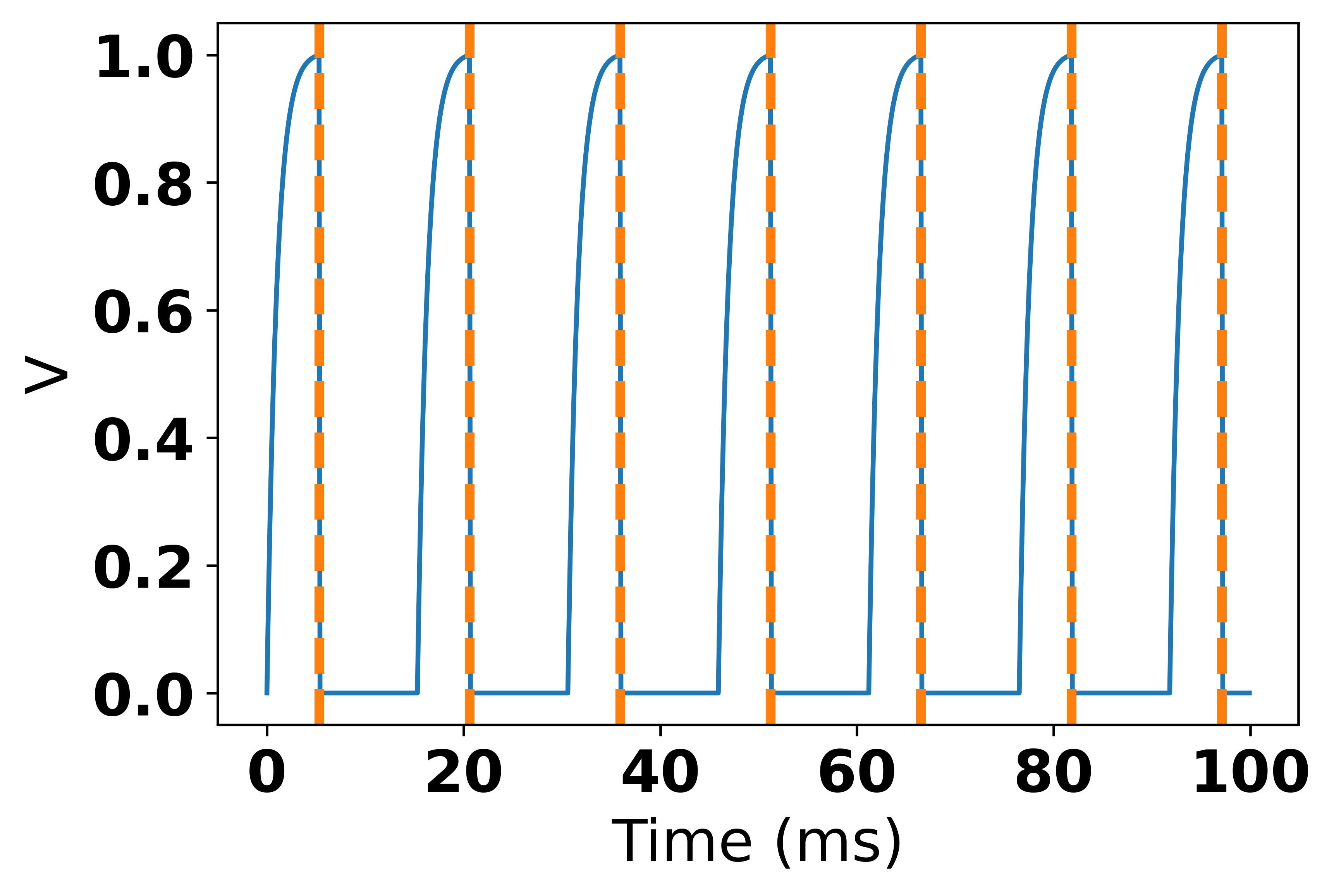}%
	}\hfill
	\subfloat[]{%
		\includegraphics[width=0.5\linewidth]{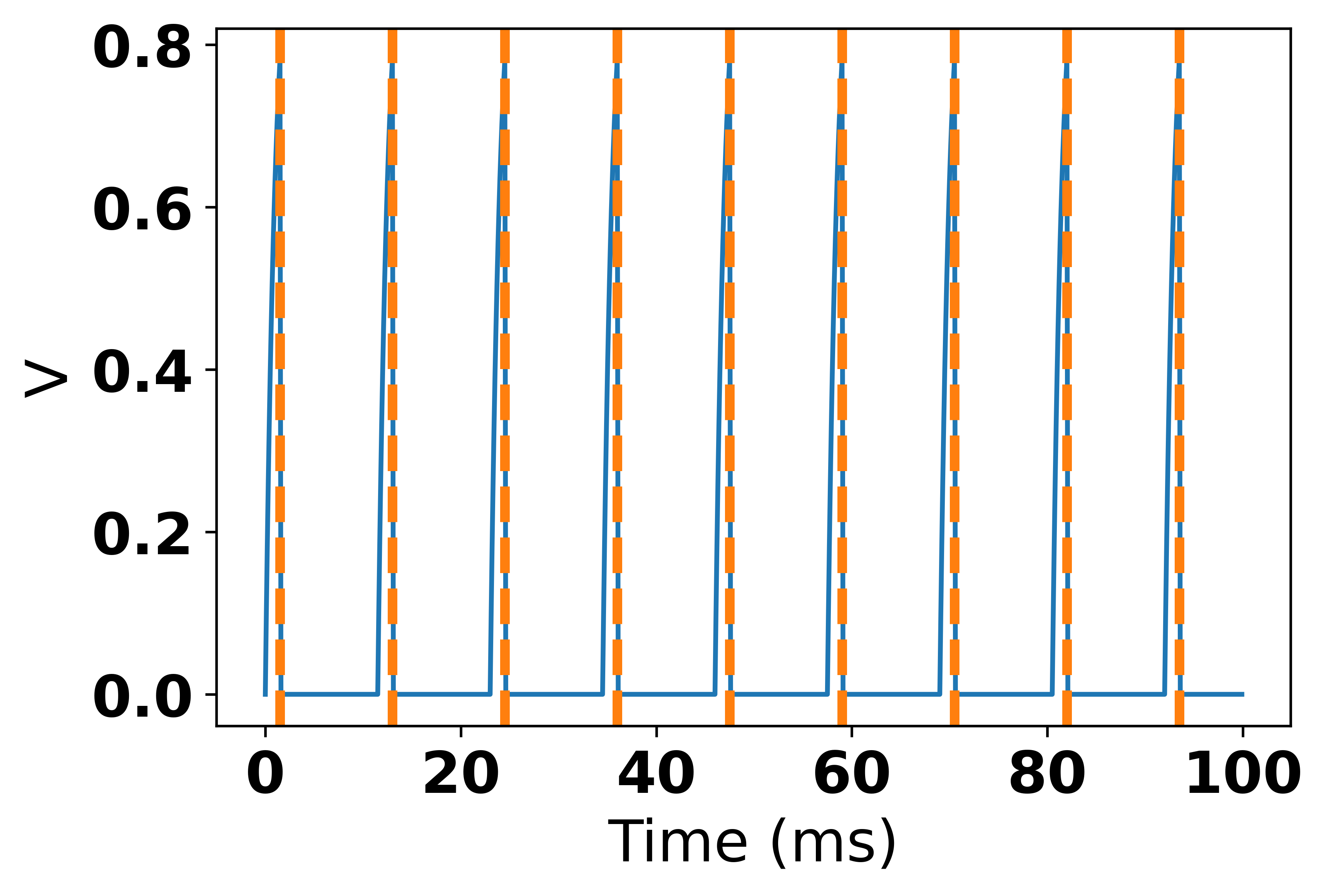}%
	}
\captionof{figure}{Spiking demonstration of the LIF with multiple threshold levels. Membrane potential is represented in blue and spike emissions are represented in orange dashed lines. a) Spiking activity at 1$v$ threshold ; b) Spiking activity at 0.8$v$ threshold }
\label{ThreshDemo}
\end{center}
\end{figure}

\begin{equation}
	\label{EqIP}
	v_{thr}(t)=\begin{cases}
		v_{thr}(t-1) + N\theta_{pos}v_{init},\,s(t-1)=1\\
		v_{thr}(t-1) - N\theta_{neg}v_{init},\,\text{otherwise}
	\end{cases}
\end{equation}

In equation \ref{EqIP}, the current voltage threshold $v_{thr}(t)$ is dependent on the previous time step's spike event. If the inter-spike-interval (ISI) \cite{Gerstner2002a} is $1ms$ , then threshold is increased by a factor of the initial threshold of the neuron $v_{init}$. The factor is decided by multiplying learning rate $\theta_{pos}$ by number of neurons in the hidden layer $N$. If a spike did not occur, the threshold is reduced by a learning rate of $\theta_{neg}$  increasing the spiking probability of the neuron. Both learning rates are small allowing IP to take place throughout the training period avoiding early saturation.
 
In order to adjust $\theta_{pos}$ and $\theta_{neg}$ (i.e., the learning rates controlling the threshold), we formulated a guided method with two dependent parameters namely information entropy \cite{Shannon1948} and active neurons\cite{Lazar2007}. We hypothesized the classifier to make less errors when the uncertainty is at lowest, given that maximum number of neurons were active(see Figure \ref{tuningcurve}). In information theory, entropy is a measure of uncertainty or average amount of information stored in a random variable \cite{Shannon1948}, whereas high number of active neurons contributes towards better pattern separation\cite{Lazar2007}. Entropy  was calculated based on the spiking probability distribution produced during training. This method is different to SpiKL-IP tuning method introduced by Zhang and Li \cite{Zhang2019} where output spiking is adjusted using reciprocals of LIF membrane time constant and resistance to produce an exponential firing distribution. Our method is more inline with \cite{Lazar2007} and \cite{Li2018} using direct threshold adjustments.

\begin{equation}
	\label{EqEntro}
	H(X)= - \sum_{i=1}^{n}P(x_{i})\log_{b}P(x_{i})
\end{equation}

For the entropy equation \ref{EqEntro}, we calculated the probability of spiking rates $P(x_{i})$, according to spiking rate probability density function (PDF) of the entire network. This PDF was obtained after a training batch is processed. In other words, our method evaluates the PDF of the entire neuron population to make changes to the excitability of all neurons. This consideration of the global activity is influenced by underlying local modifications of STDP.

\subsubsection{Neuron Pruning} 
\label{Pruning}
Neuron pruning is found to be activity dependent in biological systems \cite{Navlakha2018} and, considered as an integral part of neuronal development \cite{Yamaguchi2015}. Iglasias and Villa investigated apoptosis in a large scale ASNN trained using STDP \cite{Iglesias2006}. This study reported network stabilization and emergence of cell assemblies producing recurrent spiking sequences. The decision for apoptosis was based on neuronal firing rate. 

We adopted a similar approach where the decision to prune a neuron was based on spiking rate, calculated after completing the training cycle (i.e., STDP+IP only). With the introduction of IP which is found to be capable of learning independent features from the input \cite{Savin2010}, we hypothesized the spiking rate of a given neuron(i.e., after training) to be proportional to it's contribution towards pattern separation. It is important to note that, computational implementation of this method is a deactivation rather than a complete pruning. The computational cost of this approach is lesser compared to individual synaptic pruning, which requires checking each synaptic weight against a threshold \cite{Rathi2019,Shi2019}. Moreover, this method does not require the classifier to be trained since the pruning is solely based on spiking activity rather than network performance.

\begin{equation}
	C(x) = \sum_{i=0}^{n}\llbracket s_{i} = x \rrbracket \text{  } if\; x = 1 
	\label{EqProbCount}
\end{equation}

\begin{equation}
	P(x) = {\dfrac{C(x)}{n}}
	\label{EqProb}
\end{equation}

Considering $x$ to be a neuron in the hidden layer, our algorithm maintains a count of firing during the training period $C(x)$. In the equation, $n$ is the final time point of training simulation. If the spiking rate is higher than a threshold decided based on the firing rate distribution, it is preserved while the rest is pruned.  

\subsection{Classifier learning}
\label{classifier}
The classifier of the network is inspired by \cite{Kasabov2013} and, follows a semi-supervised approach for sample classification. Every training sample is represented by an output neuron which is evolved at each propagation. All the hidden layer neurons are connected to this output layer via excitatory synapses where ranker order \cite{Thorpe1998} rule is used for weight initiation and, spike counting is used to update the weight.

\begin{equation}
	\label{EqClsIni}
	w_{i,j}(inital) = \alpha.mod^{order(i,j)}
	\end{equation}
\begin{equation}
	\label{EqClsFi}
	w_{i,j}(final) = w_{i,j}(inital)+\sum_{t=1}^{t=n} d
\end{equation}

When $i$ pre-synaptic neuron is connected to $j$ post-synaptic neuron, the weights are initiated according to the exponent of the modulation factor $mod$. The learning parameter is denoted by $\alpha$. For the first spike arrival $order(i,j)$ is 0, therefore gets the highest weight and other weights allocated in a decreasing manner. Following the weight initiation, a drifts parameter $d$ is used to increase or decrease the initial weight according to spike arrival at each time step $t$. At the inference stage, the similarities between neurons evolved are calculated using euclidean distance to label the testing sample. 

\begin{figure*}
	\includegraphics[width=1\linewidth]{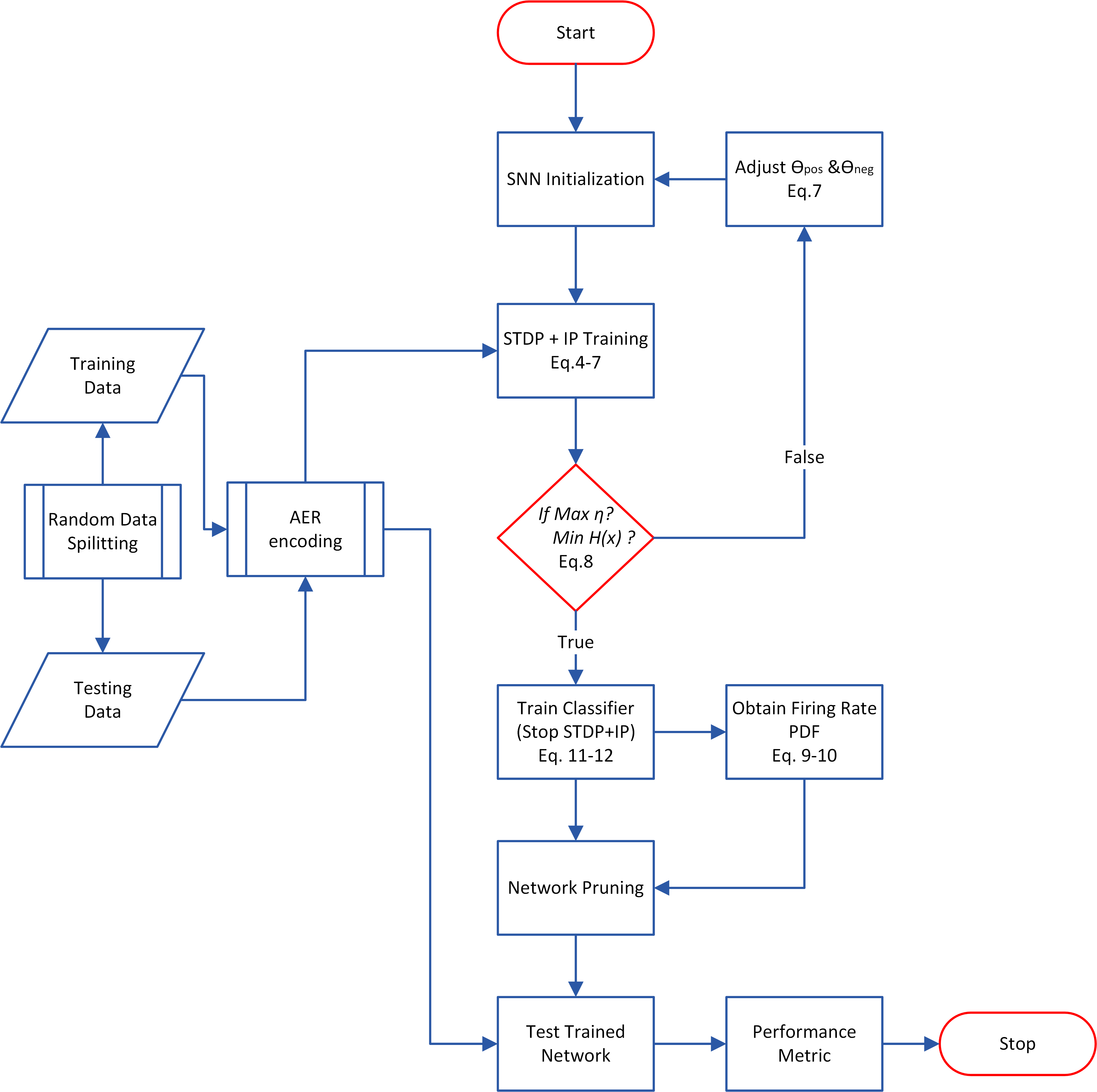}
\captionof{figure}{Flow chart representation of the ensembled learning and pruning process. Analogue EEG signals are converted to spikes using Address Event Representation(AER) \cite{Delbruck2007} encoding method \ref{AER}. STDP (Equations \ref{EqApos}, \ref{EqAneg} \& \ref{EqSynW}) and IP (Equations \ref{EqIP}) implemented as an ensemble learning method to operate simulatnoeusly. If average active neurons $\eta$ is maximized, entropy of the spiking PDF is evaluated (Equation \ref{EqEntro}) to adjust IP learning rates, $\theta_{pos}$ and $\theta_{neg}$. Once the classifier is trained (Equations \ref{EqClsIni} \& \ref{EqClsFi}), network pruning is applied (Equations \ref{EqProbCount} \& \ref{EqProb}). }
\end{figure*}

\subsection{Experimental Framework} 
\subsubsection{Data sets used }
To test the methods discussed, we used electroencephalogram(EEG) data  in a classification setup. We selected two datasets to cover EEG signals related motor movements and human emotions. For both datasets, samples were stored in comma separated values (csv) files. The format of a file is represented by an $[m,n]$ matrix where $m$ and $n$ refer to EEG channels and time points  ($t  \varepsilon  \mathbb{N}$) respectively. Each value in data represents the instantaneous voltage ($x_t  \varepsilon  \mathbb{Q}$) recorded in micro-volts. 
The first data set used was from a study conducted previously to test feasibility of SNNs in recognizing wrist positions \cite{Taylor2014}. EEG was collected from three healthy participants performing wrist flexion, extension or rest. Each position was held on command for two seconds with the participants' eyes closed to collect EEG which were sampled at 128Hz latter on. A 14 channel EEG cap was used to collect data from locations defined under international 10-20 locations. No additional artifact removal or filtering process was carried out. The final dataset consisted of 60 samples in a 14 by 128 matrix with 20 samples for each class (i.e. wrist flexion, extension or rest). This data set will be referred to as the Wrist dataset hereon.

The second dataset is widely used for emotion recognition tasks, commonly referred to as the DEAP dataset \cite{Koelstra2012}. The DEAP data set consists of EEG samples collected from 32 participants while watching 40, one minute video clips. A 32 channel EEG cap was used with electrodes placed  according to international 10-20 locations. Each participant rated the video on scales of valance, arousal, dominance and liking using a self-assessment manikin. We used the pre-processed data set which is free of artifacts caused by eye movement, filtered using 4Hz to 45Hz bandpass filter and downsampled to 128Hz. We removed six non-EEG data fields and applied an averaging window  on the time vector as a data reduction process to expedite data processing during experimentation \cite{Golmohammadi2019}. The window size was set to 32 which enabled single experiment step of 5-fold cross-validation and 70/30 split testing to complete within 30 minutes (Experiments were conducted in a core i7 machine with 16GB RAM). Final pre-processed dataset  consisted of 1086 EEG samples labeled into to two classes: $Low\:arousal = arousal\leqslant5 $  and  $high\:arousal = arousal>5 $. Each class consisted of 543 samples in a 32 by 252 matrix.

\subsubsection{Data encoding}  
EEG is a time varying analogue signal where instantaneous voltage amplitude is recorded. Since artificial spiking neurons are restricted to spiking or non-spiking states only (i.e., one or zero computational representation), the input voltage amplitudes of EEG signals should be converted to a spiking equivalent. In this study we used Address Event Representation(AER) encoding since it operates based on the rate of voltage change with a high processing efficiency \cite{Delbruck2007}. The rate of voltage change is a strong indication of neuronal events and, user defined threshold in AER allows sensitivity adjustments to such rate changes. This threshold was set to 0.5 according to the findings of our previous study \cite{Weerasinghe2021}.       

\subsubsection{Performance evaluation criteria}
\label{EvalMeth}
To evaluate the performance of each method, we selected three metrics namely: classification accuracy, F1-score, and Cohen's Kappa. We evaluated results using these metrics under 70/30 training testing split and five-fold cross-validation (CV). Before training the network, data is split (i.e. 70\% for training and 30\% testing) and, CV implemented using the training set only to analyze the training data-fit. We selected five-fold cross validation considering the size of the training data set. After CV, the network is reinitialized and, retrained with training data followed by testing with 30\% of unseen data. 
Apart from the standard accuracy measurement, we included F1-score to represent precision and recall of the classifier and, Kappa static to compare performance against random chance. We tested each method for both pattern separation capability(i.e., Robustness) and model performance under changing input and/or output environments(i.e., Adaptability) \cite{Navlakha2018}. The SNN introduced here learns through one pass learning and, we have run each simulation 30 times with randomization incorporated in sample selection, weight initiation and input mapping. This challenges the model at each train-test cycle and, would help assess the adaptability of the algorithm.   
   
\subsubsection{Experiment layout}
Experiments were conducted under three comparable approaches. Firstly, both datasets were modeled using only STDP. In the second approach, IP was applied with STDP (i.e., ensemble learning). Thirdly, a model trained with ensemble learning was pruned, based on firing rates of the neurons of the hidden layer. The initial networks consisted of 200 and 300 hidden neurons for Wrist and DEAP datasets respectively. We changed the initial network sizes due to the difference in sample sizes of the two datasets.  
Throughout the experimentation, non of the hyper-parameters (i.e. set before training) were changed, enabling a fair comparison between the approaches experimented. Only the IP learning rates were tuned to regulate spiking and, pruning rate thresholds were changed to explore the effects comparatively. The IP learning rates were restricted between $1$ to $1\times10^{-10}$ after initial experimentation. These smaller values enable the network to produce balanced firing during the entire training period \cite{Lazar2007}. The pruning thresholds were decided based on activation levels of clusters of neurons observed. All hyper-parameter values of the ASNN can be found in \ref{Hyperparameter}.  

\section{Results} 
\label{Section 4}
\subsection{Wrist dataset}
In this section we present the results obtained from different approaches experimented: STDP only, ensemble learning (STDP+IP) and, network trained with ensemble learning pruned. The raster plots and firing distributions presented were obtained after propagating the training data set through trained networks. 
  
\subsubsection{STDP and IP combined learning}
The key requirement for IP learning is the selection of values for $\theta_{pos}$ and $\theta_{neg}$ (According to the IP equations \ref{EqIP}, here we have made reference to both values in terms of magnitude only). In general, if $\theta_{pos}$ increased above 0.001 with $\theta_{neg}$ is set to zero, the spikes vanishes towards the end of the training cycle (i.e, when 70\% of data is used for training). If the conditions were reversed, the network became over-activated (See Fig.\ref{12a}). 

\begin{figure}[H]
	\begin{center}		
	\subfloat[\label{Fig.4a}]{%
		\includegraphics[width=1\linewidth]{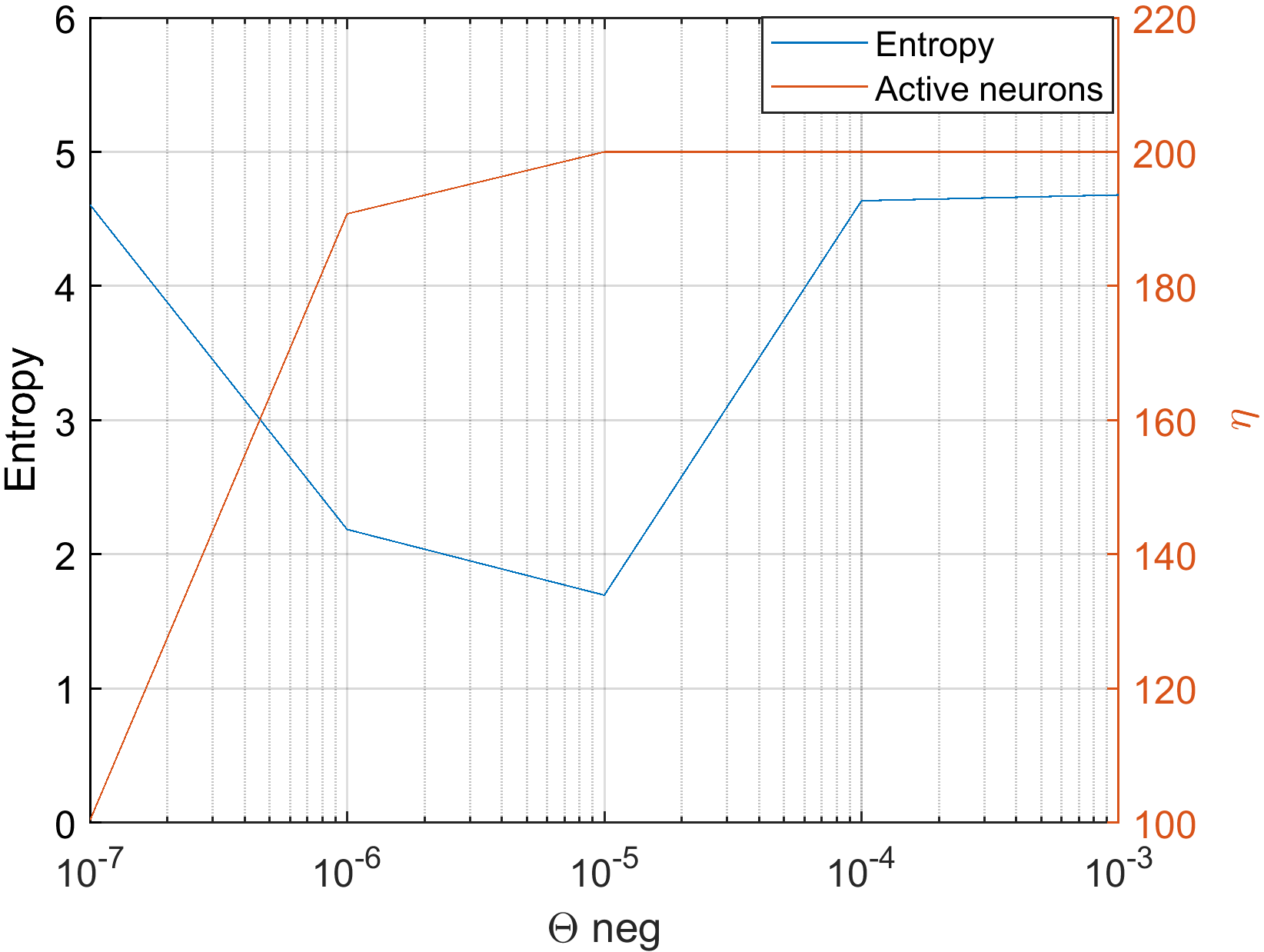}%
	}\hfill
	\subfloat[]{%
		\includegraphics[width=1\linewidth]{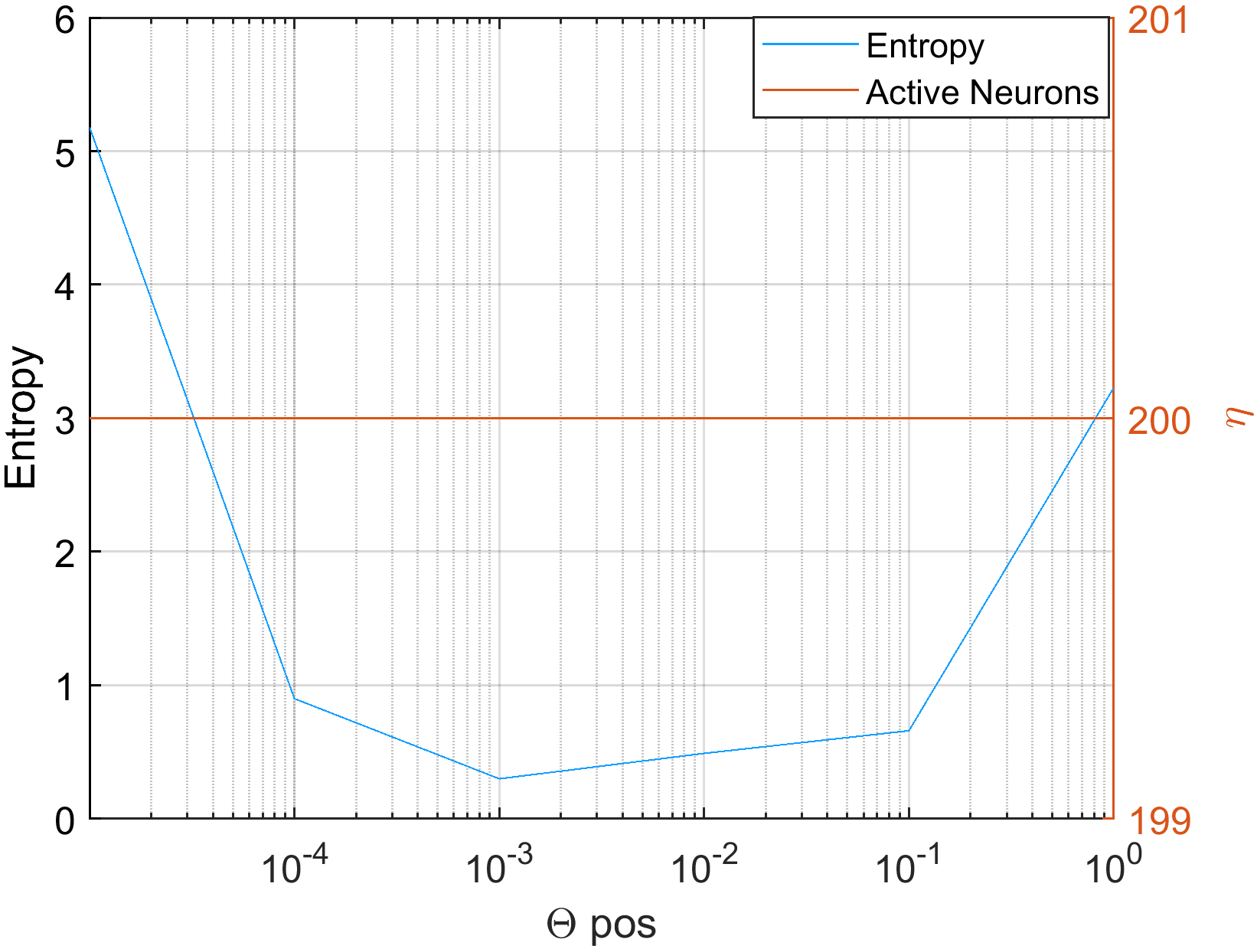}%
	}
\captionof{figure}{a) Result of sweeping $\theta_{neg}$ from  $1\times10^{-7}$ to $1\times10^{-3}$ on $\eta$ (Orange) and Entropy (Blue) while $\theta_{pos}$ is kept constant at $1\times10^{-3}$ b)Result of sweeping $\theta_{pos}$ from  $1\times10^{-5}$ to $1$ on $\eta$ (Orange) and Entropy (Blue) while $\theta_{neg}$ is kept constant at $1\times10^{-3}$  }
\label{tuningcurve}
	\end{center}
\end{figure}

As shown in Fig.\ref{Fig.4a} when $\theta_{pos}$ \& $\theta_{neg}$ was set to $1\times10^{-3}$ and  $1\times10^{-5}$ respectively, the ASNN produced minimum uncertainty under all neurons being activated at the hidden layer as per our hypothesis. We used the same heuristic method for tuning IP learning rates for both data sets(See \ref{Hyperparameter} for learning rates and other hyperparameter values).
 %However, a similar condition occurred when both $\theta_{pos}$ and  $\theta_{neg}$ were at $1\times10^{-3}$: with sparse spiking (See \ref{overActive} Fig.A.12b). Although this instance adhered to the hypothesis condition, it did not indicate spiking dynamics as seen in Fig.5b. Therefore, we selected $\theta_{pos}$ to be $1\times10^{-3}$ and $\theta_{neg}$ to be $1\times10^{-5}$. We also noted the over-activation, if $\theta_{pos}$ becomes smaller than $\theta_{neg}$(See \ref{overActive} Fig.A.12a) which was expected since there were more non-spiking instances than spiking. 

\begin{figure*}[]
	\begin{center}
	\subfloat[\label{5a}]{%
		\includegraphics[width=0.46\linewidth]{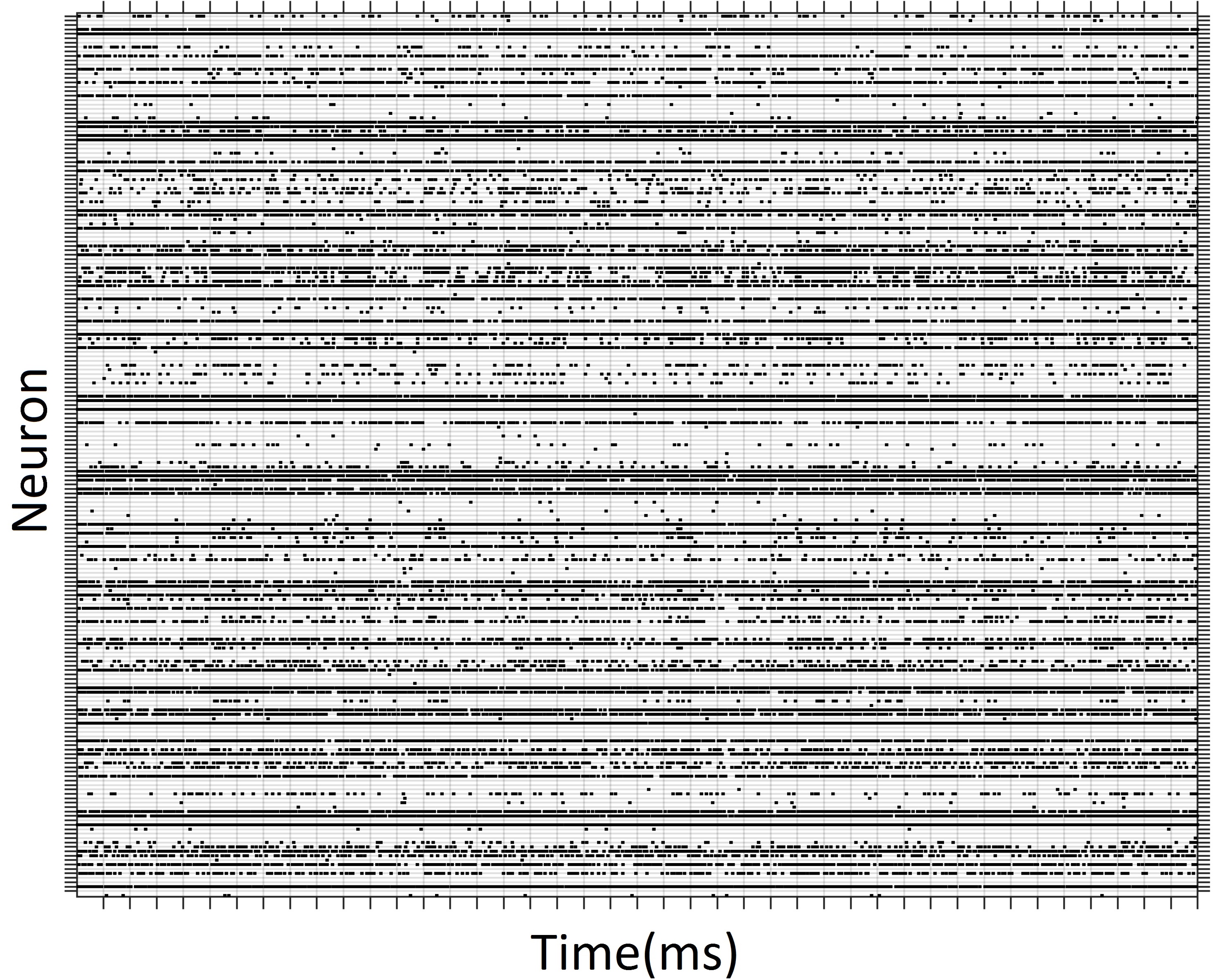}%
	}\hfill
	\subfloat[\label{5b}]{%
		\includegraphics[width=0.48\linewidth]{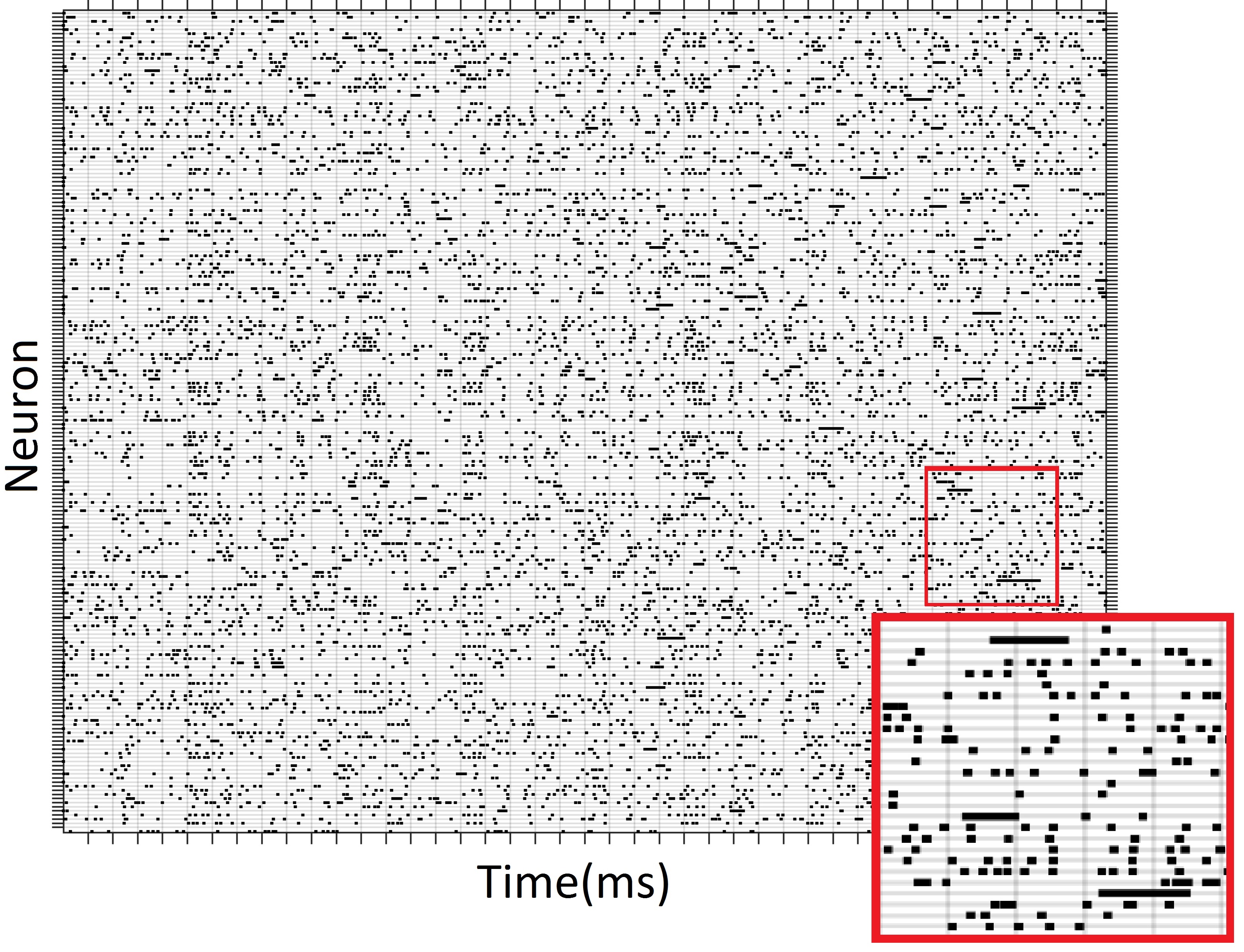}%
	}\hfill
	\subfloat[\label{5c}]{%
		\includegraphics[width=0.5\linewidth]{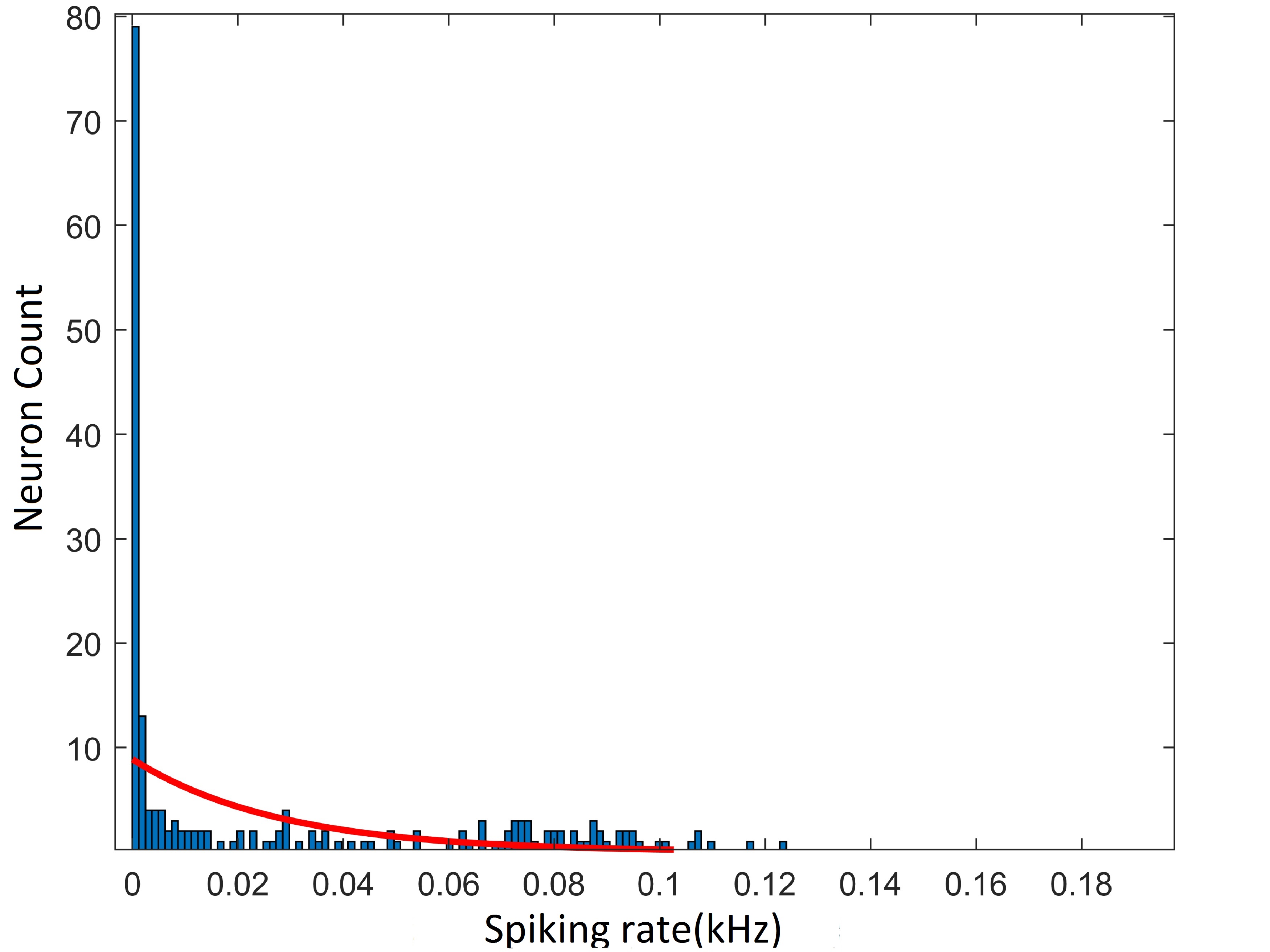}%
	}\hfill
	\subfloat[\label{5d}]{%
		\includegraphics[width=0.5\linewidth]{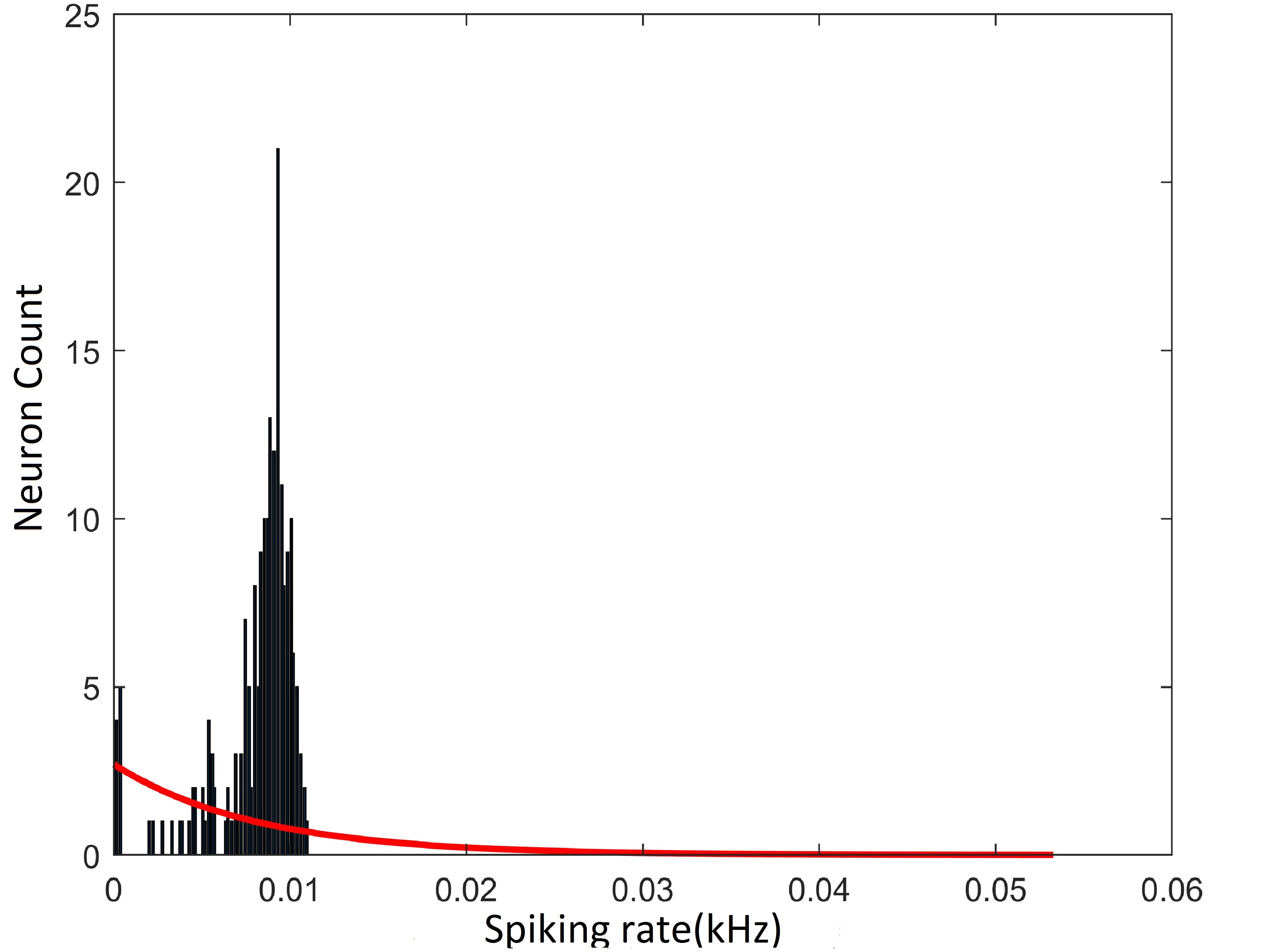}%
	}
\captionof{figure}{Spiking raster plots and firing distributions obtained after propagating the training data over the trained network a) Standalone STDP raster b) STDP+IP raster c) Standalone STDP firing distribution d) STDP+IP firing distribution}
\label{Fig5}
\end{center}
\end{figure*}

When comparing the spike train differences between Fig.\ref{5a} and Fig.\ref{5b}, sparseness of the latter is evident. This sparseness is observed both temporally(i.e., along the time axis) and spatially(i.e., among neighboring neurons). Interestingly, within the sparsity observed in Fig.\ref{5b}, scattered bursts of spiking emerged in each neuron as seen in the zoom enlargement.% Presumably, these observations represent the sensitivity of the hidden layer neurons to incoming spiking patterns based on salience. 
Standalone STDP only seemed to operate the hidden layer neurons in either firing with high or low frequency regimes, with occasional neurons maintaining firing rates in between as per Fig.\ref{5c}. This indicates high number of neurons showing low spiking rates with peripheral spiking rates being scattered fitting in to a decaying exponential. In contrast, Fig.\ref{5d} shows a left tailed normal distribution with ensemble learning. 

In terms of spiking rates, STDP recorded $0.032 (\pm0.003, range=0.01)$, whereas ensemble learning recorded  $0.009 (\pm2\times10^{-4}, range=0.001)$ over 30 cycles of random testing conducted. This translates to better efficiency in terms of information encoding with ensemble learning: which utilizes 3.6 times lesser amount of spikes on average for information encoding compared to standalone STDP $(twosample\; t-test, p < 0.05)$.  

\begin{figure*}
	\includegraphics[width=1\linewidth]{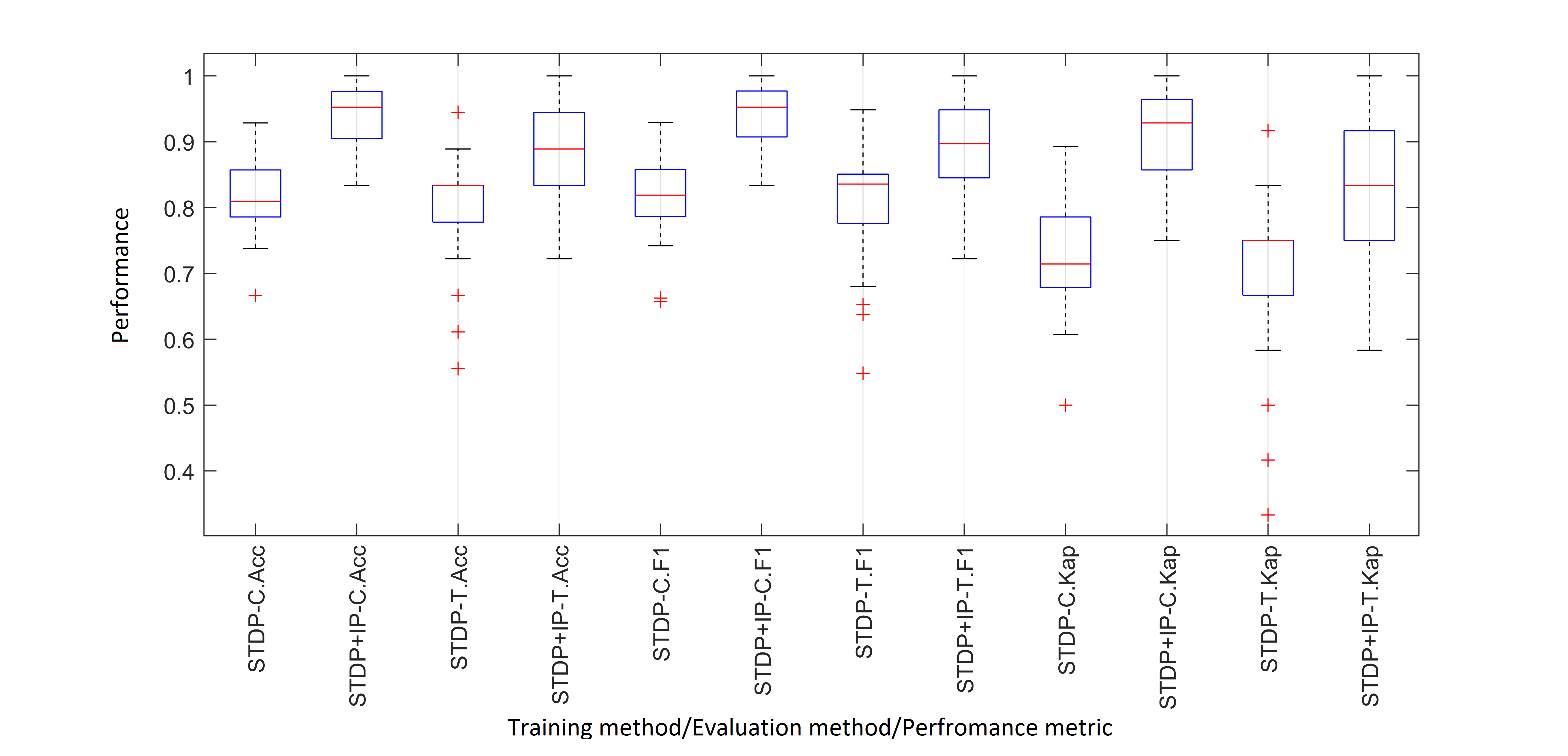}
\captionof{figure}{Perfomance comparison of STDP only and STDP+IP training. Cross validation is denoted by \emph{C} and split testing by \emph{T}. \emph{Acc} for Accuracy, \emph{F1} for F1-Score and \emph{Kap} for Kappa value.}
\label{Fig6}
\end{figure*}

Before pruning was applied, we tested the methods of STDP only and ensemble learning using the classifier presented in \ref{classifier}. We ran 5-fold cross-validation and a 70/30 split testing across the Accuracy, F1-Score and Kappa matrices. The ensemble learning demonstrated better robustness with increased average performance and lesser outliers compared to standalone STDP as per Fig.\ref{Fig6}. However, when considering the results of cross-validation and split testing in ensemble learning showed a slight over-fitting.

%Lesser variation under cross-validation and testing indicates the adaptive capability of STDP+IP over standalone STDP. However, the over-fitting is resulted due to the tendency of the SNN to learn minute features in the data. This raises the question as to which neurons in the network are responsible for the over-fit.

\subsubsection{Network Pruning}
Based on the pruning hypothesis in \ref{Pruning}, we selected three pruning rate thresholds from the network firing rate distribution. This selection enabled us to  analyze network performance comparatively. The pruning spike-rate thresholds were, 0.017 , 0.052 and 0.087 which pruned on average ~17\% , ~21\% and ~24\% of the network respectively. These three thresholds represent the lowest clusters of contribution according to our pruning hypothesis.

\begin{figure}[]
	\begin{center}
	\subfloat[\label{7a}]{%
		\includegraphics[width=1\linewidth]{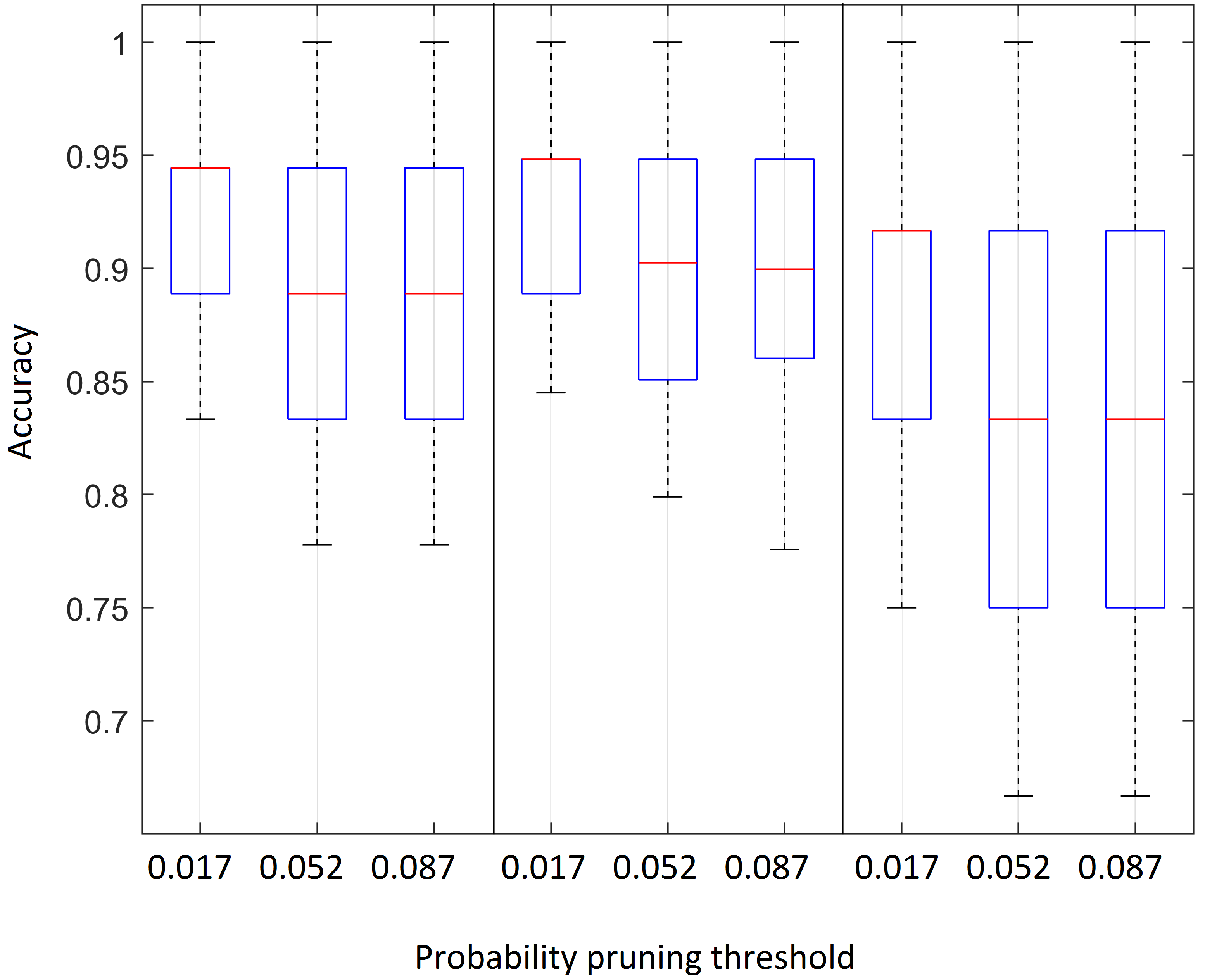}%
	}\hfill
	\subfloat[\label{7b}]{%
		\includegraphics[width=1\linewidth]{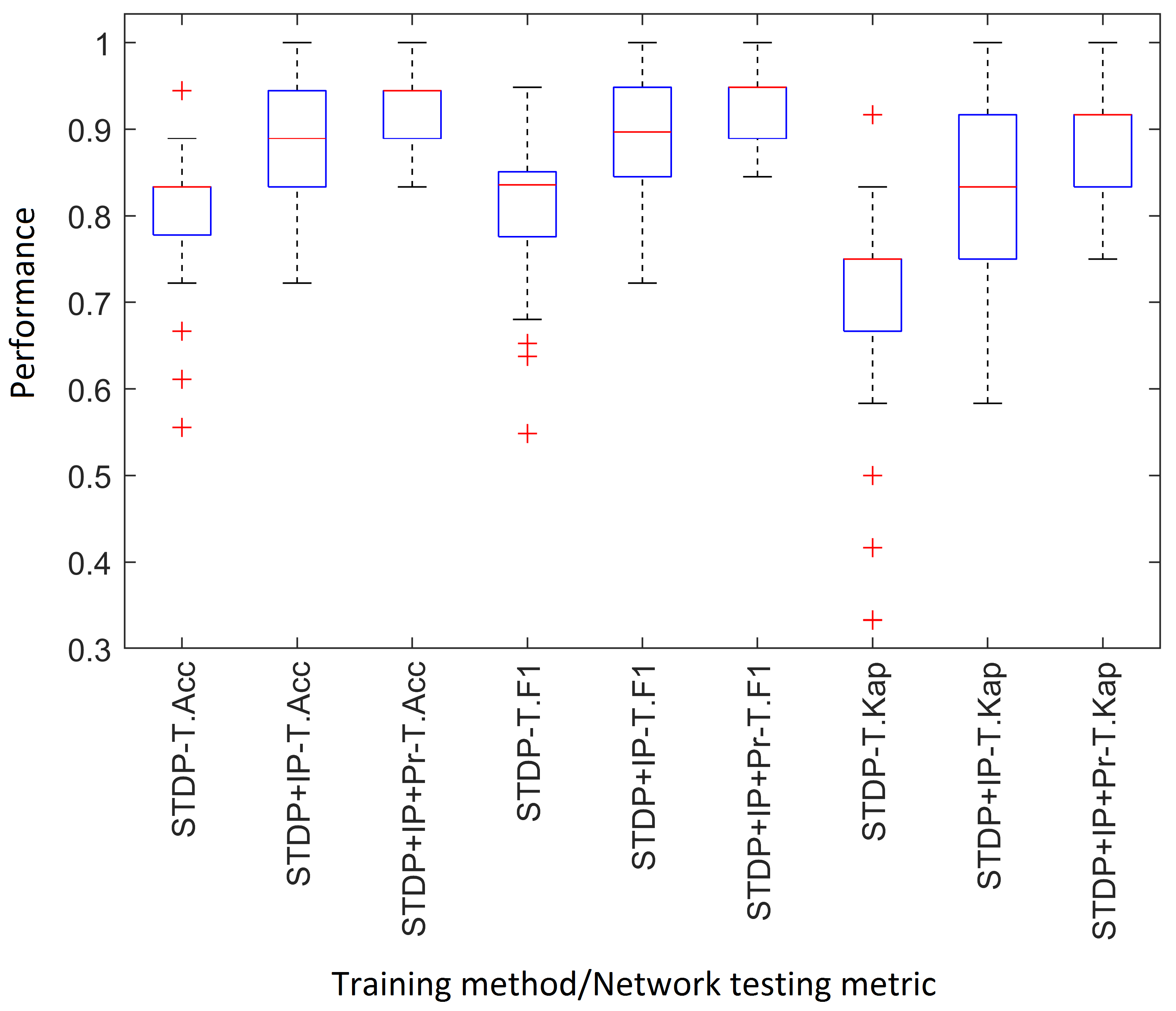}%
	}
\captionof{figure}{a) Pruned network accuracy performance compared under 70/30 train-test split b) Comparison of 70/30 train-split performance after STDP only, STDP+IP and STDP+IP Pruned (denoted as \emph{Pr}) approaches. Split testing is denoted by \emph{T},\emph{Acc} for Accuracy, \emph{F1} for F1-Score and \emph{Kap} for Kappa value.}	
\end{center}
\end{figure}

From the three pruning thresholds, the lowest threshold resulted the highest, whereas the others yielded lower performances as shown in Fig.\ref{7a}: in terms of robustness and adaptability. This pattern was observed across all performance metrics. In particular, the higher thresholds dropped performance by ~5\% on average. 

Interestingly, the pruned SNN (i.e., under the lowest threshold) demonstrated better performance comparison to standalone STDP and STDP+IP approaches as per Fig.\ref{7b}. 
%This implies the pruned network to have better generalization capability compared to other approaches. Referring to the over-fit observed with STDP+IP, it is highly probable that the cluster of low firing neurons to cause the over-fitting condition.    

Table.\ref{Wrist-Table} summarizes the average robustness resulted during different approaches discussed in this section. An over-fit of ~1-2\% is present in standalone STDP which had grown to an over-fit of ~4-8\% with STDP+IP across all matrices. Pruning results indicate the countering effect that has elevated generalization capability by ~2-5\%.     

\begin{table}[]
	\begin{threeparttable}

	\caption{Average ML performances summarized for Wrist data}
	\label{Wrist-Table}
	\centering
	\begin{tabular*}{\columnwidth}{c
			*{2}{S[table-format=1.2]}
			S[table-format=2.2] 
			*{3}{S[table-format=1.2]}
			S[table-format=2.0]
		}
		\toprule
		\multicolumn{1}{c}{Metric}
		& \multicolumn{2}{c}{STDP only}& 
		& \multicolumn{2}{c}{\makecell{STDP + IP}}
		& \multicolumn{1}{c}{Pruned\tnote{*}} \\
		\cmidrule{2-3}
		\cmidrule{5-6}
		 & {C.V} & {Test} & & {C.V} & {Test} & & \\
		\midrule
		Accuracy   & 0.814 & 0.796 &  & 0.936  & 0.880  & 0.917\\
		F1-Score   & 0.818 & 0.807 &  & 0.937  & 0.889  & 0.924\\
		Kappa	   & 0.721 & 0.694 &  & 0.904  & 0.819  & 0.875\\

		\bottomrule
	
	\end{tabular*}
	\footnotesize
	\begin{tablenotes}[para,flushleft]
		\item[*]{Testing results after ~17\% of the original network was pruned following STDP+IP training}
	
	\end{tablenotes}
	\end{threeparttable}
\end{table}

\subsection{DEAP data set}
This section presents the results of performing the same experiments as detailed for the Wrist dataset with the DEAP dataset. 

\subsubsection{STDP and IP combined learning}
\begin{figure}
	\includegraphics[width=1\linewidth]{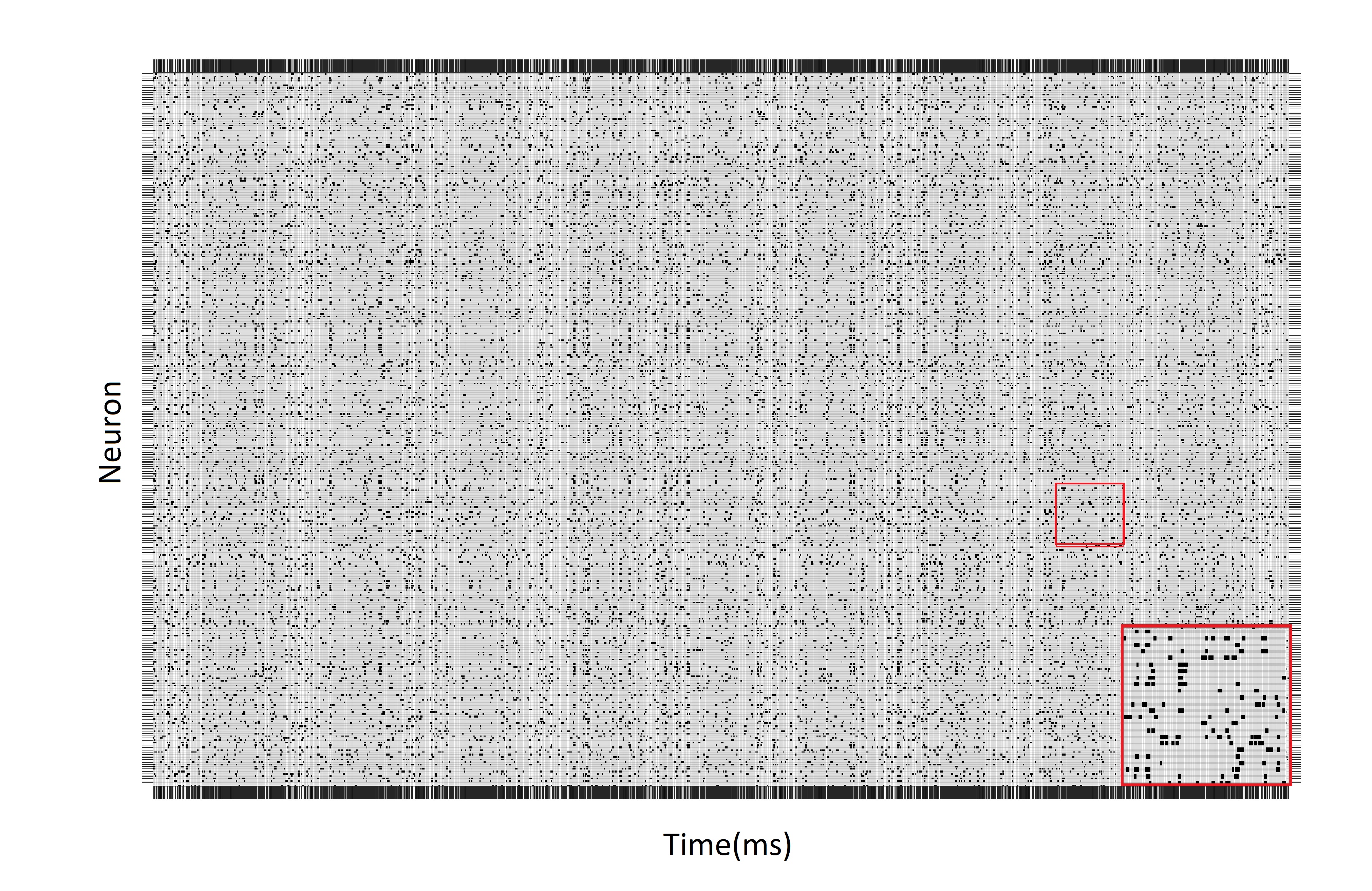}
	\captionof{figure}{Spiking raster plot obtained after propagating the training data over the trained network using STDP+IP}
	\label{DEAPIP}
\end{figure}

With the DEAP dataset, we observed sparseness of spiking with ensemble learning, when compared with standalone STDP. Fig.\ref{DEAPIP} illustrates sparse spiking after ensemble learning. 

The zoom enlargement in Fig.\ref{DEAPIP} shows the sizes of spike bursts observed. Two observation can be stated when these bursts are compared with that of Wrist data experiment: a) bursts observed are much shorter, b) diversity of burst sizes are minimal. Moreover, observing the entire raster plot, similar size bursts are aligned vertically indicating minimum temporal diversity. In contrary, the network trained with STDP only, showed large group of neurons firing constantly at high rates.      

\begin{figure}[]
	\begin{center}
		\subfloat[\label{9a}]{%
		\includegraphics[width=1\linewidth]{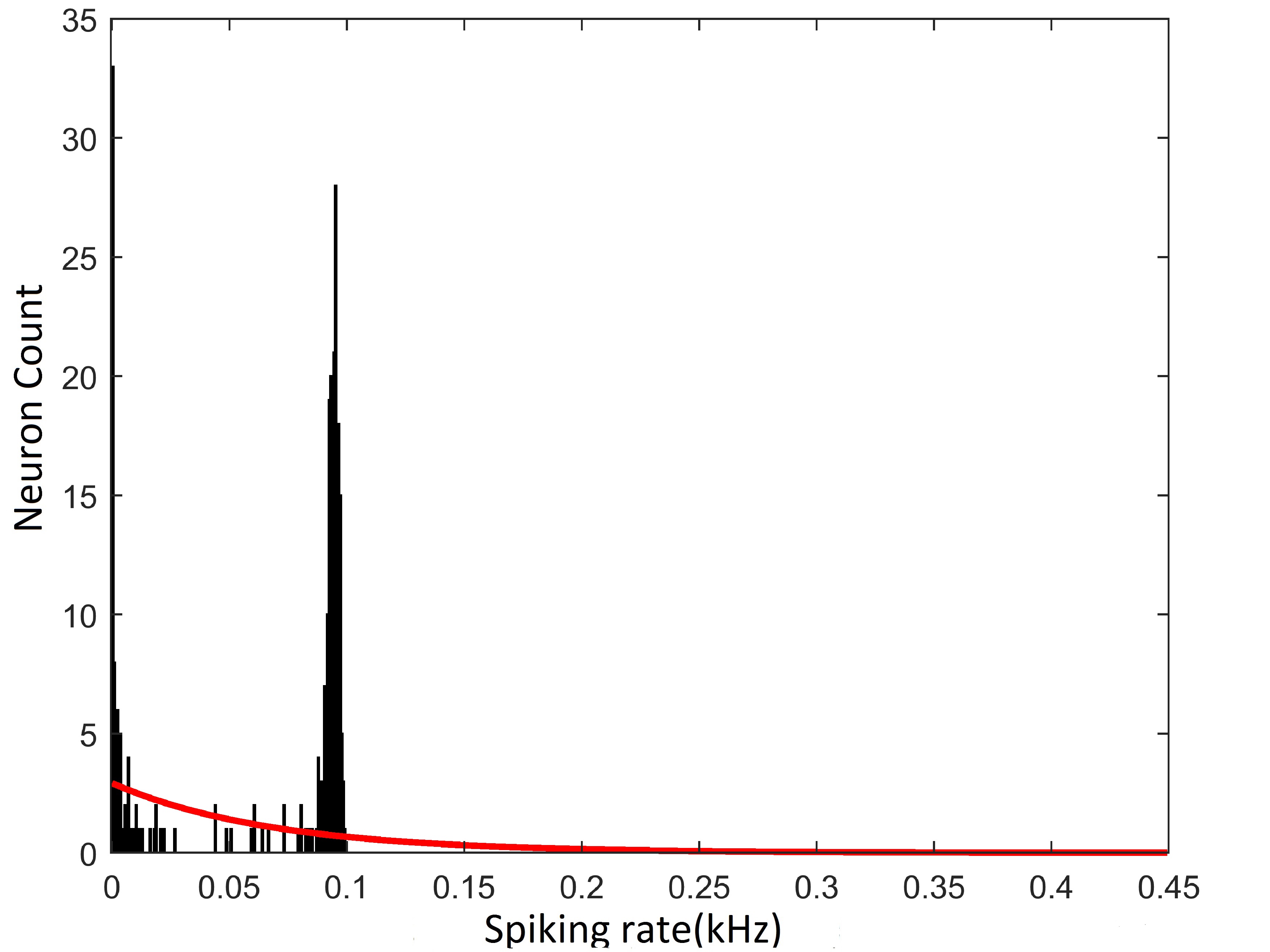}%
	}\hfill
	\subfloat[\label{9b}]{%
		\includegraphics[width=1\linewidth]{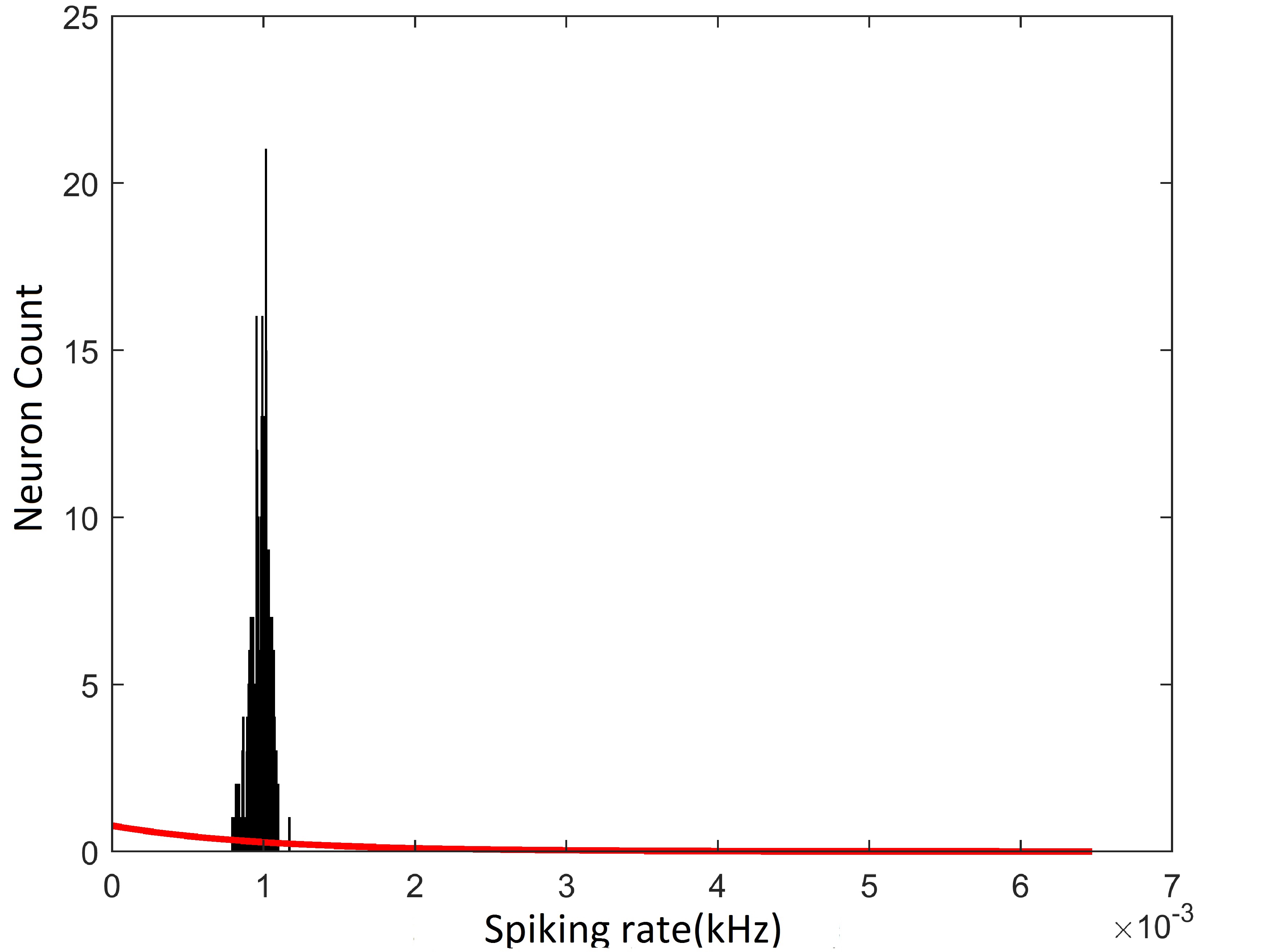}%
	}
	\captionof{figure}{a) Spiking distribution under standalone STDP b) Spiking distribution under STDP+IP }
	
\end{center}
\end{figure}

The spiking distributions in STDP produced a bimodal distribution where large number of neurons were concentrated on high or low firing rates as per Fig.\ref{9a}. After applying IP, this firing rate diversity was further reduced and, resulted in a normal spike distribution (see Fig.\ref{9b}). When these spiking distributions are compared with Wrist data distributions, the lack of diversity in spiking rates are evident. 
%Presumably, this can be caused by the lack of diversity in the input stimuli.

The spiking rates of STDP recorded $0.066 (\pm0.002, range=0.009)$, whereas ensemble learning recorded  $0.001 (\pm3.05\times10^{-7}, range=1.7\times10^{-6})$ over 30 cycles of random testing conducted. This translates to better efficiency in terms of information encoding with ensemble learning: which utilizes 65 times lesser amount of spikes on average for information encoding compared to standalone STDP $(twosample t-test, p < 0.05)$.  

\begin{figure*}
	\includegraphics[width=1\linewidth]{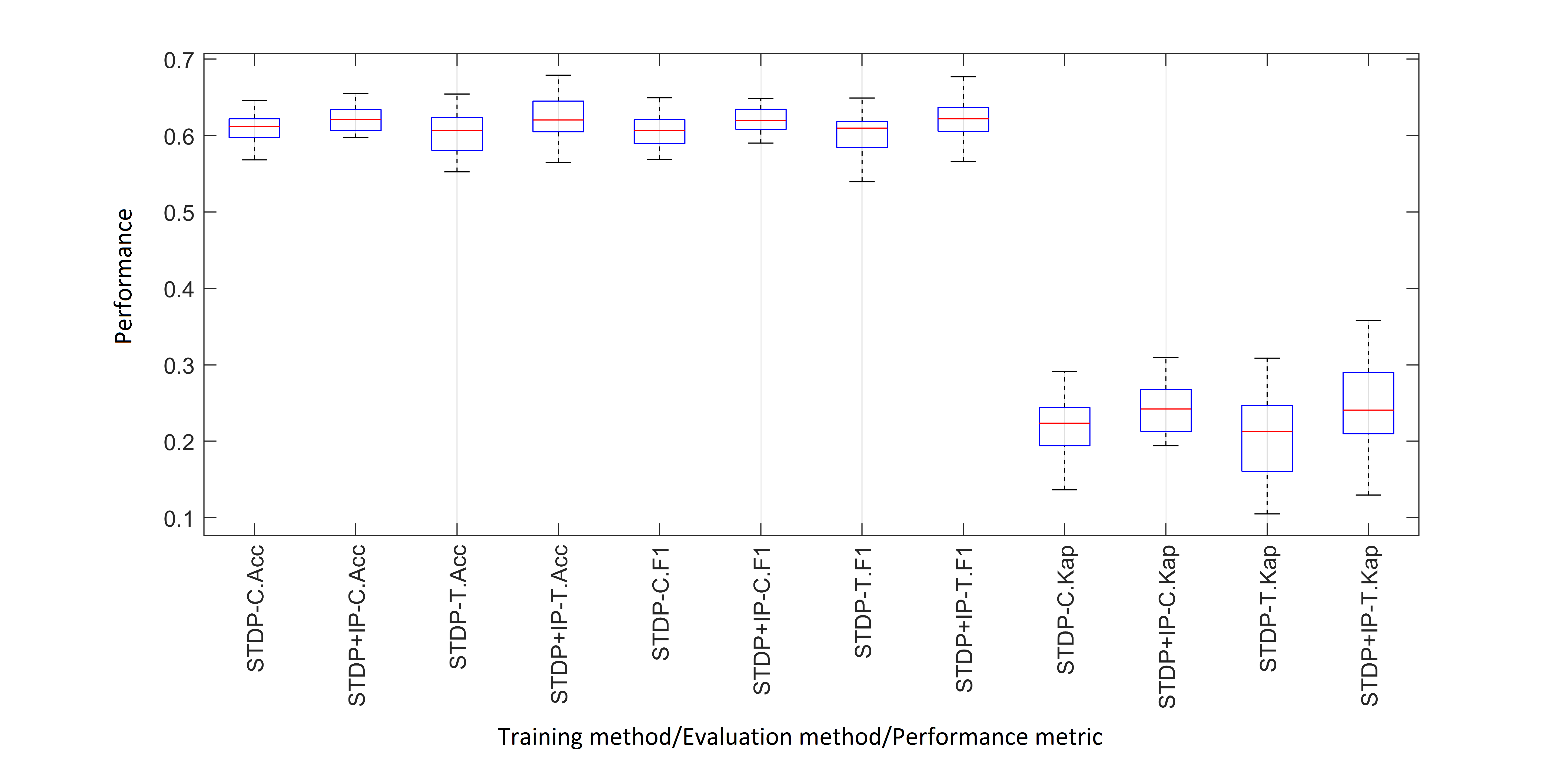}
	\captionof{figure}{Perfomance comparison of STDP only and STDP+IP training for the DEAP dataset. Cross validation is denoted by \emph{C} and split testing by \emph{T}. \emph{Acc} for Accuracy, \emph{F1} for F1-Score and \emph{Kap} for Kappa value.}
	\label{Fig10}
\end{figure*}

In terms of robustness, STDP+IP showed 1-3\% improvement in all matrices of measurement, than standalone STDP. Adaptability was also better with IP which is indicated by the lower dispersion of performance (Fig.\ref{Fig10}). We did not observe any over-fitting condition with the DEAP data unlike with the Wrist data, but the overall performance in terms of pattern separation was low.

\subsubsection{Network Pruning}
\begin{figure}
	\includegraphics[width=1\linewidth]{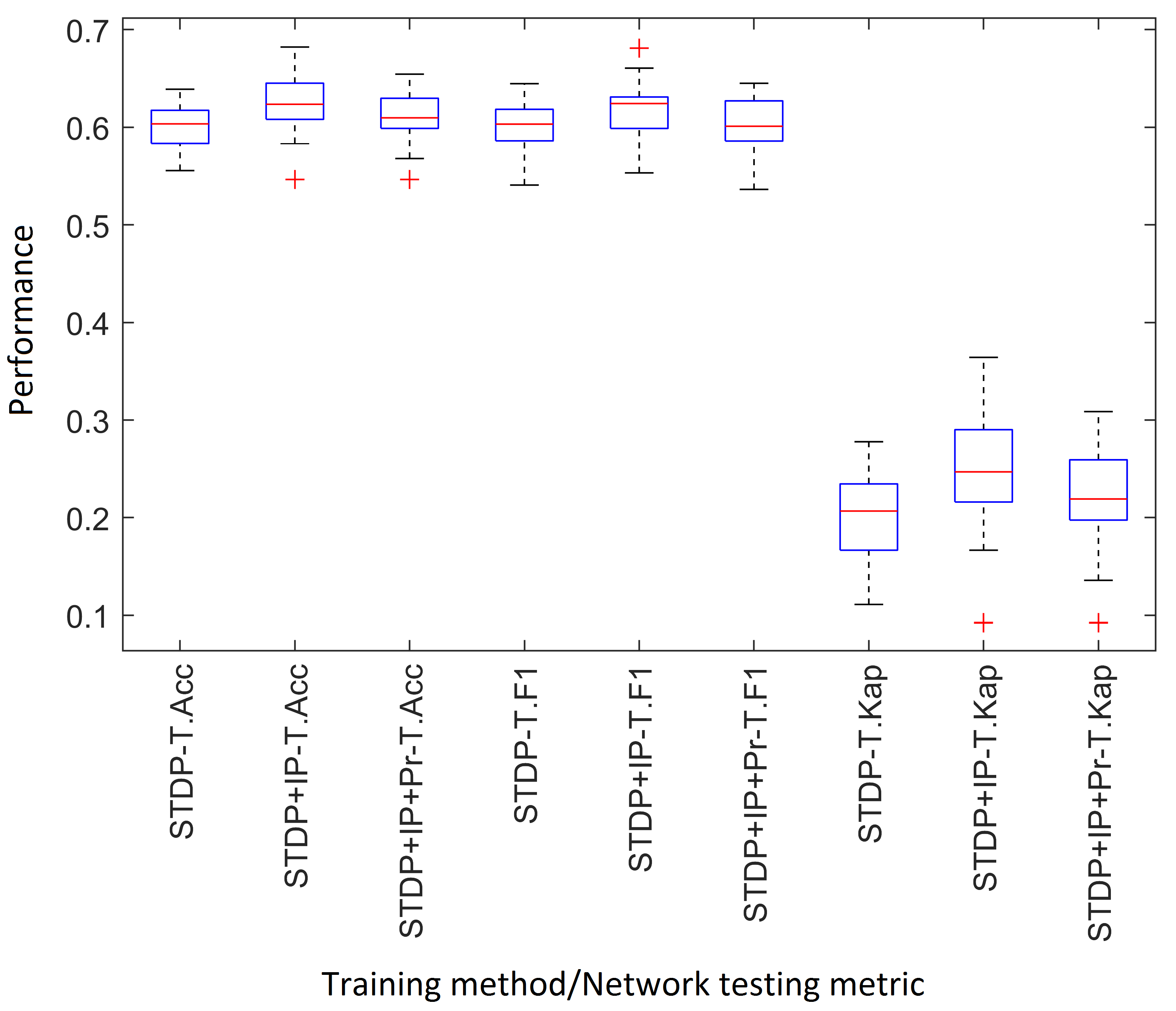}
	\captionof{figure}{ Comparison of 70/30 train-split performance under different approaches of STDP only, STDP+IP and STDP+IP Pruned (denoted as \emph{Pr}). Split testing is denoted by \emph{T},\emph{Acc} for Accuracy, \emph{F1} for F1-Score and \emph{Kap} for Kappa value.}
\end{figure}

We pruned the network trained using ensemble learning under three different pruning spike-rate thresholds of $8.2\times10^{-4}$,$8.8\times10^{-4}$ and $9.3\times10^{-4}$. These thresholds were assigned based on the three lowest spiking rates observed in the network. The average testing accuracy was recorded at $0.62(\pm0.034, range = 0.11)$, $0.61(\pm0.034, range = 0.12)$ and $0.61(\pm0.031, range = 0.16)$ for $8.2\times10^{-4}$, $8.8\times10^{-4}$ and $9.3\times10^{-4}$ pruning thresholds respectively. These thresholds pruned the network on average by ~26\%,~29\% and ~36\%. The initial pruning threshold, $8.2\times10^{-4}$ yeilded higher robustness, than the standalone STDP, with ~26\% of the neurons removed. The performance summary under three approaches are given in Table.\ref{DEAP-Table}.

\begin{table}[]
	\begin{threeparttable}
	\caption{Average ML performances summarized for DEAP Data}
	\label{DEAP-Table}
	\centering
	\begin{tabular*}{\columnwidth}{c
			*{2}{S[table-format=1.2]}
			S[table-format=2.2] 
			*{3}{S[table-format=1.2]}
			S[table-format=2.0]
		}
		\toprule
		\multicolumn{1}{c}{Metric}
		& \multicolumn{2}{c}{STDP only}& 
		& \multicolumn{2}{c}{STDP+IP}
		& \multicolumn{1}{c}{Pruned\tnote{*}} \\
		\cmidrule{2-3}
		\cmidrule{5-6}
		& {C.V} & {Test} & & {C.V} & {Test} & & \\
		\midrule
		Accuracy   & 0.606 & 0.601 &  & 0.622  & 0.624  & 0.617\\
		F1-Score   & 0.605 & 0.597 &  & 0.617  & 0.618  & 0.612\\
		Kappa	   & 0.212 & 0.202 &  & 0.243  & 0.248  & 0.234\\
		
		\bottomrule
	\end{tabular*}
	\footnotesize
	\begin{tablenotes}[para,flushleft]
		\item[*]{Testing results after ~26\% of the original network was pruned following STDP+IP training}
	
	\end{tablenotes}
	\end{threeparttable}
\end{table}
  
\section{Discussion}
\label{Section 5}
The results presented clearly indicates the superiority of the ensemble learning approach over standalone STDP, in terms of robustness, adaptability and, processing efficiency (See Fig.\ref{Fig6} and Fig.\ref{Fig10}). Moreover, the application of neuron pruning after ensemble learning countered over-fitting conditions (i.e., in Wrist data) or gradually decreased inference performance. Interestingly, even if ~25\% the network was pruned, the average inference performance remained better than standalone STDP for both data sets.

After applying ensemble learning, we observed a cluster of low firing neurons separated from the main firing distribution with Wrist data (see Fig.\ref{5d}). This was not the case with DEAP data (see Fig.\ref{9b}) where diversity of spiking rates observed was minimum. The pruning of low firing neurons rectified over-fitting in Wrist data and reduced the performance with DEAP data. In computational biology, Savin and Joshi proposed IP and synaptic scaling to complement STDP to enable single neurons to learn independent components from the input \cite{Savin2010}. Accordingly, we propose the low firing neuron cluster to represent minute features causing over-fitting. Furthermore, the similarity of the input spikes may have caused less diversity in spiking distribution that resulted in lower accuracy with DEAP data.

The novel IP tuning method introduced in this paper use two parameters to regulate spiking activity, namely; neuron activation and reduced entropy (refer section \ref{IP}). These parameters are measured using the entire network activity, enabling \emph{low but not too low} level of spiking \cite{Li2013} and, maintaining a firing homeostasis while preventing runaway synaptic potentiation (see sparser firing in Fig.\ref{5a} \& Fig.\ref{DEAPIP}). In ML perspective, runaway synaptic potentiation decreases efficiency due to overpopulation of spikes. These findings keeps inline with previous studies  \cite{Chen2013,Watt2010}, which discussed the ability of IP in countering runway synaptic potentiation.

Our method of IP adaptation is different compared to SPIKL-IP \cite{Zhang2019}, which uses a singular objective of information maximization without STDP.  According to information maximization rule presented by Bell and Sejnowski \cite{Bell1995}, maximizing output entropy would increase the mutual transfer of information from input to output. They also pointed out, that an exponential output firing distribution would correspond to a maximum entropy. Interestingly, when comparing the performances(See Fig.\ref{Fig6} and Fig.\ref{Fig10}) and firing distributions(See Fig.\ref{5d} \& Fig.\ref{9a}) of Wrist and DEAP data, we can observe the following: if standalone STDP produces an exponential firing distribution (i.e. maximum information transfer)(See Fig.\ref{Fig5}), proposed ensemble method would greatly enhance performance(~8-13\% see Table \ref{Wrist-Table}). However, if standalone STDP does not produce an exponential output firing distribution (See DEAP output firing distribution Fig.\ref{9b}), the performance enhancement with ensemble learning is less (~1-3\% see Table\ref{DEAP-Table}). This translates to either STDP learning rates not enabling maximum information transfer or inadequacy of salient information from the input. Nevertheless, in both instances of modeling discussed in this paper, ensemble learning produced normal firing distributions. Therefore, according to our experiments, tuning learning rates to obtain a desired exponential \cite{Zhang2019} or Weibull \cite{Li2013} firing output distribution may not be essential when IP is applied with STDP in a multi-spike environment for better ML performance. 

The method of network adaptability discussed here differs from previous studies \cite{Rathi2019,Shi2019} due to the use of ensemble plasticity and, spike-rate neuron pruning used instead of synaptic pruning. Since the neuron spiking is more representative of the temporal features, this method can amalgamate a soft-pruning mechanism \cite{Shi2019} without removing the neuron completely which can be beneficial if the robustness of the network is lower as per the instance with the DEAP data. Furthermore, since this adaptive mechanism is based on unsupervised learning, it can be used as a Neural Architecture Search (NAS) \cite{Mellor2020} before complete training or  hyperparameter optimization. The combined method of ensemble learning and pruning can also be converted to online learning since IP learning rates and pruning thresholds are adjusted primarily based on spiking activity, rather than final inference result.  

An important spiking behavior was observed which appears to push neurons to operate at a critical stage or ``edge of chaos''\cite{Bertschinger2004} which helps achieve higher performance and computational efficiency\cite{Li2017}. These bursts of spikes that we observed  (see Fig.\ref{5b}) are often referred to as avalanches. High performance at the presence of avalanches is clearly evident when spiking behavior of ensemble learning is compared between Wrist data and DEAP data. This performance enhancement ability of avalanches has been proposed previously in biological modeling \cite{Li2018}. This is an important area that needs further investigations that may open windows for ASNN to perform better since the information representation capability can be further enhanced beyond spike time, rate or rank.      

\section{Conclusion} 
\label{Section 6}
In this work we present an unsupervised ensemble learning method (by combining STDP and IP) with a spike regulation strategy, developed based on information theory.  This method is then amalgamated with a biologically inspired network pruning mechanism. We have evaluated these methods on a rigorous ML framework and, attempted to interpret results at the fundamental level of spikes. The spike-based ensemble learning and pruning methods discussed in this paper can enhance robustness and increase efficiency particularly in multi-spike learning tasks. These methods can be adapted for batch-learning, online-learning or architectural searches before extensive training and/or hyper-parameter optimizations.

In terms of limitations, the data sets we have used in this study are balanced, which restrict us from discussing its performance on unbalanced data. Moreover, we used a common spike encoding mechanism to both tasks; this is a crucial point in modeling with obvious implications on ML performance. Apart from addressing the said limitations, we are motivated to: a) explore further on ASNNs operating in ``edge of chaos'', guided by ensemble learning and apoptosis b) develop online ASNN algorithms using methods discussed c) implement participant specific ASNN models for neuropsychological hypothesis generation, as apart of our future work.


\nocite{*}
\bibliographystyle{elsarticle-num} 

\bibliography{Chapter6Neurocomputing,1.Growing Pruning ANNs,1.Paper2DEAP,Affective Computing,Brain and Signals,Chapter 4,EEG,Emotion Recognition,IEEE Access,NeuCube,Neuromorphic,Neuron Models,Pruning and Growin Neurons-Computer,SNN Advantages,SNNs & EEG,SNNs,SNNs-Temporal Aspect,SNNs-Unsup Learning,Spatiotemporal Data,STDP-Unsup,}

\begin{thebibliography}{10}
\expandafter\ifx\csname url\endcsname\relax
  \def\url#1{\texttt{#1}}\fi
\expandafter\ifx\csname urlprefix\endcsname\relax\def\urlprefix{URL }\fi
\expandafter\ifx\csname href\endcsname\relax
  \def\href#1#2{#2} \def\path#1{#1}\fi

\bibitem{Pfeiffer2018}
M.~Pfeiffer, T.~Pfeil,
  \href{https://www.frontiersin.org/article/10.3389/fnins.2018.00774/full}{{Deep
  Learning With Spiking Neurons: Opportunities and Challenges}}, Frontiers in
  Neuroscience 12~(October) (oct 2018).
\newblock \href {https://doi.org/10.3389/fnins.2018.00774}
  {\path{doi:10.3389/fnins.2018.00774}}.
\newline\urlprefix\url{https://www.frontiersin.org/article/10.3389/fnins.2018.00774/full}

\bibitem{Maass1997a}
W.~Maass,
  \href{https://linkinghub.elsevier.com/retrieve/pii/S0893608097000117}{{Networks
  of spiking neurons: The third generation of neural network models}}, Neural
  Networks 10~(9) (1997) 1659--1671.
\newblock \href {https://doi.org/10.1016/S0893-6080(97)00011-7}
  {\path{doi:10.1016/S0893-6080(97)00011-7}}.
\newline\urlprefix\url{https://linkinghub.elsevier.com/retrieve/pii/S0893608097000117}

\bibitem{Tavanaei2019a}
A.~Tavanaei, M.~Ghodrati, S.~R. Kheradpisheh, T.~Masquelier, A.~Maida,
  \href{https://linkinghub.elsevier.com/retrieve/pii/S0893608018303332}{{Deep
  learning in spiking neural networks}}, Neural Networks 111~(January 2019)
  (2019) 47--63.
\newblock \href {http://arxiv.org/abs/2546440} {\path{arXiv:2546440}}, \href
  {https://doi.org/10.1016/j.neunet.2018.12.002}
  {\path{doi:10.1016/j.neunet.2018.12.002}}.
\newline\urlprefix\url{https://linkinghub.elsevier.com/retrieve/pii/S0893608018303332}

\bibitem{Roy2019}
K.~Roy, A.~Jaiswal, P.~Panda, \href{http://dx.doi.org/10.1038/s41586-019-1677-2
  http://www.nature.com/articles/s41586-019-1677-2}{{Towards spike-based
  machine intelligence with neuromorphic computing}}, Nature 575~(7784) (2019)
  607--617.
\newblock \href {https://doi.org/10.1038/s41586-019-1677-2}
  {\path{doi:10.1038/s41586-019-1677-2}}.
\newline\urlprefix\url{http://dx.doi.org/10.1038/s41586-019-1677-2
  http://www.nature.com/articles/s41586-019-1677-2}

\bibitem{Greff2017}
K.~Greff, R.~K. Srivastava, J.~Koutnik, B.~R. Steunebrink, J.~Schmidhuber,
  \href{http://ieeexplore.ieee.org/document/7508408/}{{LSTM: A Search Space
  Odyssey}}, IEEE Transactions on Neural Networks and Learning Systems 28~(10)
  (2017) 2222--2232.
\newblock \href {http://arxiv.org/abs/1503.04069} {\path{arXiv:1503.04069}},
  \href {https://doi.org/10.1109/TNNLS.2016.2582924}
  {\path{doi:10.1109/TNNLS.2016.2582924}}.
\newline\urlprefix\url{http://ieeexplore.ieee.org/document/7508408/}

\bibitem{Kasabov2013}
N.~Kasabov, K.~Dhoble, N.~Nuntalid, G.~Indiveri,
  \href{https://linkinghub.elsevier.com/retrieve/pii/S0893608012003139}{{Dynamic
  evolving spiking neural networks for on-line spatio- and spectro-temporal
  pattern recognition}}, Neural Networks 41 (2013) 188--201.
\newblock \href {https://doi.org/10.1016/j.neunet.2012.11.014}
  {\path{doi:10.1016/j.neunet.2012.11.014}}.
\newline\urlprefix\url{https://linkinghub.elsevier.com/retrieve/pii/S0893608012003139}

\bibitem{Dora2016}
S.~Dora, K.~Subramanian, S.~Suresh, N.~Sundararajan,
  \href{https://linkinghub.elsevier.com/retrieve/pii/S0925231215010942}{{Development
  of a Self-Regulating Evolving Spiking Neural Network for classification
  problem}}, Neurocomputing 171 (2016) 1216--1229.
\newblock \href {https://doi.org/10.1016/j.neucom.2015.07.086}
  {\path{doi:10.1016/j.neucom.2015.07.086}}.
\newline\urlprefix\url{https://linkinghub.elsevier.com/retrieve/pii/S0925231215010942}

\bibitem{Bi1998}
G.-q. Bi, M.-m. Poo,
  \href{https://www.jneurosci.org/content/jneuro/18/24/10464.full.pdf
  http://www.jneurosci.org/lookup/doi/10.1523/JNEUROSCI.18-24-10464.1998
  https://www.jneurosci.org/lookup/doi/10.1523/JNEUROSCI.18-24-10464.1998}{{Synaptic
  Modifications in Cultured Hippocampal Neurons: Dependence on Spike Timing,
  Synaptic Strength, and Postsynaptic Cell Type}}, The Journal of Neuroscience
  18~(24) (1998) 10464--10472.
\newblock \href {https://doi.org/10.1523/JNEUROSCI.18-24-10464.1998}
  {\path{doi:10.1523/JNEUROSCI.18-24-10464.1998}}.
\newline\urlprefix\url{https://www.jneurosci.org/content/jneuro/18/24/10464.full.pdf
  http://www.jneurosci.org/lookup/doi/10.1523/JNEUROSCI.18-24-10464.1998
  https://www.jneurosci.org/lookup/doi/10.1523/JNEUROSCI.18-24-10464.1998}

\bibitem{Song2000}
S.~Song, K.~D. Miller, L.~F. Abbott, \href{http://neurosci.nature.com
  http://www.nature.com/articles/nn0900{\_}919}{{Competitive Hebbian learning
  through spike-timing-dependent synaptic plasticity}}, Nature Neuroscience
  3~(9) (2000) 919--926.
\newblock \href {https://doi.org/10.1038/78829} {\path{doi:10.1038/78829}}.
\newline\urlprefix\url{http://neurosci.nature.com
  http://www.nature.com/articles/nn0900{\_}919}

\bibitem{Abbott2000a}
L.~F. Abbott, S.~B. Nelson,
  \href{http://www.nature.com/articles/nn1100{\_}1178}{{Synaptic plasticity:
  taming the beast}}, Nature Neuroscience 3~(S11) (2000) 1178--1183.
\newblock \href {https://doi.org/10.1038/81453} {\path{doi:10.1038/81453}}.
\newline\urlprefix\url{http://www.nature.com/articles/nn1100{\_}1178}

\bibitem{Desai1999}
N.~S. Desai, L.~C. Rutherford, G.~G. Turrigiano,
  \href{https://www.nature.com/articles/nn0699{\_}515}{{Plasticity in the
  intrinsic excitability of cortical pyramidal neurons}}, Nature Neuroscience
  2~(6) (1999) 515--520.
\newblock \href {https://doi.org/10.1038/9165} {\path{doi:10.1038/9165}}.
\newline\urlprefix\url{https://www.nature.com/articles/nn0699{\_}515}

\bibitem{Zhang2003}
W.~Zhang, D.~J. Linden, {The other side of the engram: Experience-driven
  changes in neuronal intrinsic excitability}, Nature Reviews Neuroscience
  4~(11) (2003) 885--900.
\newblock \href {https://doi.org/10.1038/nrn1248} {\path{doi:10.1038/nrn1248}}.

\bibitem{Frick2005}
A.~Frick, D.~Johnston,
  \href{https://onlinelibrary.wiley.com/doi/10.1002/neu.20148}{{Plasticity of
  dendritic excitability}}, Journal of Neurobiology 64~(1) (2005) 100--115.
\newblock \href {https://doi.org/10.1002/neu.20148}
  {\path{doi:10.1002/neu.20148}}.
\newline\urlprefix\url{https://onlinelibrary.wiley.com/doi/10.1002/neu.20148}

\bibitem{Diehl2015b}
P.~U. Diehl, M.~Cook,
  \href{http://journal.frontiersin.org/Article/10.3389/fncom.2015.00099/abstract}{{Unsupervised
  learning of digit recognition using spike-timing-dependent plasticity}},
  Frontiers in Computational Neuroscience 9 (aug 2015).
\newblock \href {https://doi.org/10.3389/fncom.2015.00099}
  {\path{doi:10.3389/fncom.2015.00099}}.
\newline\urlprefix\url{http://journal.frontiersin.org/Article/10.3389/fncom.2015.00099/abstract}

\bibitem{Hao2020}
Y.~Hao, X.~Huang, M.~Dong, B.~Xu, {A biologically plausible supervised learning
  method for spiking neural networks using the symmetric STDP rule}, Neural
  Networks 121 (2020) 387--395.
\newblock \href {http://arxiv.org/abs/1812.06574} {\path{arXiv:1812.06574}},
  \href {https://doi.org/10.1016/j.neunet.2019.09.007}
  {\path{doi:10.1016/j.neunet.2019.09.007}}.

\bibitem{Navlakha2018}
S.~Navlakha, Z.~Bar-Joseph, A.~L. Barth,
  \href{https://linkinghub.elsevier.com/retrieve/pii/S1364661317302000}{{Network
  Design and the Brain}}, Trends in Cognitive Sciences 22~(1) (2018) 64--78.
\newblock \href {https://doi.org/10.1016/j.tics.2017.09.012}
  {\path{doi:10.1016/j.tics.2017.09.012}}.
\newline\urlprefix\url{https://linkinghub.elsevier.com/retrieve/pii/S1364661317302000}

\bibitem{Shi2019}
Y.~Shi, L.~Nguyen, S.~Oh, X.~Liu, D.~Kuzum,
  \href{https://www.frontiersin.org/article/10.3389/fnins.2019.00405/full}{{A
  Soft-Pruning Method Applied During Training of Spiking Neural Networks for
  In-memory Computing Applications}}, Frontiers in Neuroscience 13~(APR) (2019)
  1--13.
\newblock \href {https://doi.org/10.3389/fnins.2019.00405}
  {\path{doi:10.3389/fnins.2019.00405}}.
\newline\urlprefix\url{https://www.frontiersin.org/article/10.3389/fnins.2019.00405/full}

\bibitem{Rathi2019}
N.~Rathi, P.~Panda, K.~Roy,
  \href{https://arxiv.org/ftp/arxiv/papers/1710/1710.04734.pdf
  https://ieeexplore.ieee.org/document/8325325/
  http://arxiv.org/abs/1710.04734}{{STDP-Based Pruning of Connections and
  Weight Quantization in Spiking Neural Networks for Energy-Efficient
  Recognition}}, IEEE Transactions on Computer-Aided Design of Integrated
  Circuits and Systems 38~(4) (2019) 668--677.
\newblock \href {http://arxiv.org/abs/1710.04734} {\path{arXiv:1710.04734}},
  \href {https://doi.org/10.1109/TCAD.2018.2819366}
  {\path{doi:10.1109/TCAD.2018.2819366}}.
\newline\urlprefix\url{https://arxiv.org/ftp/arxiv/papers/1710/1710.04734.pdf
  https://ieeexplore.ieee.org/document/8325325/
  http://arxiv.org/abs/1710.04734}

\bibitem{Iglesias2006}
J.~Iglesias, A.~E.~P. Villa, \href{http://inforge.unil.ch/
  http://link.springer.com/10.1007/11840817{\_}99}{{Neuronal Cell Death and
  Synaptic Pruning Driven by Spike-Timing Dependent Plasticity}}, in: Lecture
  Notes in Computer Science (including subseries Lecture Notes in Artificial
  Intelligence and Lecture Notes in Bioinformatics), Vol. 4131 LNCS, Springer,
  Berlin, Heidelberg, 2006, pp. 953--962.
\newblock \href {https://doi.org/10.1007/11840817_99}
  {\path{doi:10.1007/11840817_99}}.
\newline\urlprefix\url{http://inforge.unil.ch/
  http://link.springer.com/10.1007/11840817{\_}99}

\bibitem{Lazar2007}
A.~Lazar, G.~Pipa, J.~Triesch,
  \href{https://linkinghub.elsevier.com/retrieve/pii/S0893608007000469}{{Fading
  memory and time series prediction in recurrent networks with different forms
  of plasticity}}, Neural Networks 20~(3) (2007) 312--322.
\newblock \href {https://doi.org/10.1016/j.neunet.2007.04.020}
  {\path{doi:10.1016/j.neunet.2007.04.020}}.
\newline\urlprefix\url{https://linkinghub.elsevier.com/retrieve/pii/S0893608007000469}

\bibitem{Weerasinghe2021}
M.~M.~A. Weerasinghe, J.~I. Espinosa-Ramos, G.~Y. Wang, D.~Parry,
  \href{https://ieeexplore.ieee.org/document/9493873/}{{Incorporating
  Structural Plasticity Approaches in Spiking Neural Networks for EEG
  Modelling}}, IEEE Access 9 (2021) 117338--117348.
\newblock \href {https://doi.org/10.1109/ACCESS.2021.3099492}
  {\path{doi:10.1109/ACCESS.2021.3099492}}.
\newline\urlprefix\url{https://ieeexplore.ieee.org/document/9493873/}

\bibitem{Savin2010}
C.~Savin, P.~Joshi, J.~Triesch,
  \href{https://dx.plos.org/10.1371/journal.pcbi.1000757
  http://www.ncbi.nlm.nih.gov/pubmed/20421937
  http://www.pubmedcentral.nih.gov/articlerender.fcgi?artid=PMC2858697}{{Independent
  component analysis in spiking neurons.}}, PLoS computational biology 6~(4)
  (2010) e1000757.
\newblock \href {https://doi.org/10.1371/journal.pcbi.1000757}
  {\path{doi:10.1371/journal.pcbi.1000757}}.
\newline\urlprefix\url{https://dx.plos.org/10.1371/journal.pcbi.1000757
  http://www.ncbi.nlm.nih.gov/pubmed/20421937
  http://www.pubmedcentral.nih.gov/articlerender.fcgi?artid=PMC2858697}

\bibitem{Stemmler1999}
M.~Stemmler, C.~Koch, \href{http://www.nature.com/articles/nn0699{\_}521}{{How
  voltage-dependent conductances can adapt to maximize the information encoded
  by neuronal firing rate}}, Nature Neuroscience 2~(6) (1999) 521--527.
\newblock \href {https://doi.org/10.1038/9173} {\path{doi:10.1038/9173}}.
\newline\urlprefix\url{http://www.nature.com/articles/nn0699{\_}521}

\bibitem{Hodgkin1952}
A.~L. HODGKIN, A.~F. HUXLEY,
  \href{https://onlinelibrary.wiley.com/doi/abs/10.1113/jphysiol.1952.sp004764
  https://onlinelibrary.wiley.com/doi/10.1113/jphysiol.1952.sp004764
  http://www.ncbi.nlm.nih.gov/pubmed/12991237
  http://www.pubmedcentral.nih.gov/articlerender.fcgi?artid=PMC1392413}{{A
  quantitative description of membrane current and its application to
  conduction and excitation in nerve.}}, The Journal of physiology 117~(4)
  (1952) 500--44.
\newblock \href {https://doi.org/10.1113/jphysiol.1952.sp004764}
  {\path{doi:10.1113/jphysiol.1952.sp004764}}.
\newline\urlprefix\url{https://onlinelibrary.wiley.com/doi/abs/10.1113/jphysiol.1952.sp004764
  https://onlinelibrary.wiley.com/doi/10.1113/jphysiol.1952.sp004764
  http://www.ncbi.nlm.nih.gov/pubmed/12991237
  http://www.pubmedcentral.nih.gov/articlerender.fcgi?artid=PMC1392413}

\bibitem{Triesch2007}
J.~Triesch, \href{https://direct.mit.edu/neco/article/19/4/885-909/7172
  http://www.ncbi.nlm.nih.gov/pubmed/17348766}{{Synergies between intrinsic and
  synaptic plasticity mechanisms}}, Neural Computation 19~(4) (2007) 885--909.
\newblock \href {https://doi.org/10.1162/neco.2007.19.4.885}
  {\path{doi:10.1162/neco.2007.19.4.885}}.
\newline\urlprefix\url{https://direct.mit.edu/neco/article/19/4/885-909/7172
  http://www.ncbi.nlm.nih.gov/pubmed/17348766}

\bibitem{Li2011}
C.~Li, \href{http://ieeexplore.ieee.org/document/5873129/}{{A Model of Neuronal
  Intrinsic Plasticity}}, IEEE Transactions on Autonomous Mental Development
  3~(4) (2011) 277--284.
\newblock \href {https://doi.org/10.1109/TAMD.2011.2159379}
  {\path{doi:10.1109/TAMD.2011.2159379}}.
\newline\urlprefix\url{http://ieeexplore.ieee.org/document/5873129/}

\bibitem{Li2013}
C.~Li, Y.~Li, \href{http://ieeexplore.ieee.org/document/6257429/}{{A
  Spike-Based Model of Neuronal Intrinsic Plasticity}}, IEEE Transactions on
  Autonomous Mental Development 5~(1) (2013) 62--73.
\newblock \href {https://doi.org/10.1109/TAMD.2012.2211101}
  {\path{doi:10.1109/TAMD.2012.2211101}}.
\newline\urlprefix\url{http://ieeexplore.ieee.org/document/6257429/}

\bibitem{Li2016}
C.~Li, Y.~Li, \href{http://ieeexplore.ieee.org/document/7358047/}{{A Review on
  Synergistic Learning}}, IEEE Access 4 (2016) 119--134.
\newblock \href {https://doi.org/10.1109/ACCESS.2015.2509005}
  {\path{doi:10.1109/ACCESS.2015.2509005}}.
\newline\urlprefix\url{http://ieeexplore.ieee.org/document/7358047/}

\bibitem{Zhang2019}
W.~Zhang, P.~Li,
  \href{https://www.frontiersin.org/article/10.3389/fnins.2019.00031/full}{{Information-Theoretic
  Intrinsic Plasticity for Online Unsupervised Learning in Spiking Neural
  Networks}}, Frontiers in Neuroscience 13~(FEB) (2019) 1--14.
\newblock \href {https://doi.org/10.3389/fnins.2019.00031}
  {\path{doi:10.3389/fnins.2019.00031}}.
\newline\urlprefix\url{https://www.frontiersin.org/article/10.3389/fnins.2019.00031/full}

\bibitem{Zhang2019a}
A.~Zhang, H.~Zhou, X.~Li, W.~Zhu,
  \href{https://doi.org/10.1016/j.neucom.2019.07.009}{{Fast and robust learning
  in Spiking Feed-forward Neural Networks based on Intrinsic Plasticity
  mechanism}}, Neurocomputing 365 (2019) 102--112.
\newblock \href {https://doi.org/10.1016/j.neucom.2019.07.009}
  {\path{doi:10.1016/j.neucom.2019.07.009}}.
\newline\urlprefix\url{https://doi.org/10.1016/j.neucom.2019.07.009}

\bibitem{Zhang2021a}
W.~Zhang, P.~Li,
  \href{https://direct.mit.edu/neco/article/33/7/1886/100577/Skip-Connected-Self-Recurrent-Spiking-Neural}{{Skip-Connected
  Self-Recurrent Spiking Neural Networks With Joint Intrinsic Parameter and
  Synaptic Weight Training}}, Neural Computation 33~(7) (2021) 1886--1913.
\newblock \href {https://doi.org/10.1162/neco_a_01393}
  {\path{doi:10.1162/neco_a_01393}}.
\newline\urlprefix\url{https://direct.mit.edu/neco/article/33/7/1886/100577/Skip-Connected-Self-Recurrent-Spiking-Neural}

\bibitem{Li2018}
X.~Li, W.~Wang, F.~Xue, Y.~Song,
  \href{http://dx.doi.org/10.1016/j.physa.2017.08.053}{{Computational modeling
  of spiking neural network with learning rules from STDP and intrinsic
  plasticity}}, Physica A: Statistical Mechanics and its Applications 491
  (2018) 716--728.
\newblock \href {https://doi.org/10.1016/j.physa.2017.08.053}
  {\path{doi:10.1016/j.physa.2017.08.053}}.
\newline\urlprefix\url{http://dx.doi.org/10.1016/j.physa.2017.08.053}

\bibitem{Morris1999}
R.~G. Morris,
  \href{https://linkinghub.elsevier.com/retrieve/pii/S0361923099001823}{{D.O.
  Hebb: The Organization of Behavior, Wiley: New York; 1949}} (nov 1999).
\newblock \href {https://doi.org/10.1016/S0361-9230(99)00182-3}
  {\path{doi:10.1016/S0361-9230(99)00182-3}}.
\newline\urlprefix\url{https://linkinghub.elsevier.com/retrieve/pii/S0361923099001823}

\bibitem{Turney2012}
S.~G. Turney, J.~W. Lichtman,
  \href{https://dx.plos.org/10.1371/journal.pbio.1001352}{{Reversing the
  outcome of synapse elimination at developing neuromuscular junctions in vivo:
  Evidence for synaptic competition and its mechanism}}, PLoS Biology 10~(6)
  (2012) e1001352.
\newblock \href {https://doi.org/10.1371/journal.pbio.1001352}
  {\path{doi:10.1371/journal.pbio.1001352}}.
\newline\urlprefix\url{https://dx.plos.org/10.1371/journal.pbio.1001352}

\bibitem{Wysoski2008}
S.~G. Wysoski, L.~Benuskova, N.~Kasabov,
  \href{https://linkinghub.elsevier.com/retrieve/pii/S0925231208002191}{{Fast
  and adaptive network of spiking neurons for multi-view visual pattern
  recognition}}, Neurocomputing 71~(13-15) (2008) 2563--2575.
\newblock \href {https://doi.org/10.1016/j.neucom.2007.12.038}
  {\path{doi:10.1016/j.neucom.2007.12.038}}.
\newline\urlprefix\url{https://linkinghub.elsevier.com/retrieve/pii/S0925231208002191}

\bibitem{Wang2014}
J.~Wang, A.~Belatreche, L.~Maguire, T.~M. McGinnity,
  \href{https://linkinghub.elsevier.com/retrieve/pii/S0925231214005785}{{An
  online supervised learning method for spiking neural networks with adaptive
  structure}}, Neurocomputing 144 (2014) 526--536.
\newblock \href {https://doi.org/10.1016/j.neucom.2014.04.017}
  {\path{doi:10.1016/j.neucom.2014.04.017}}.
\newline\urlprefix\url{https://linkinghub.elsevier.com/retrieve/pii/S0925231214005785}

\bibitem{Roy2017}
S.~Roy, A.~Basu, \href{http://ieeexplore.ieee.org/document/7508492/}{{An Online
  Unsupervised Structural Plasticity Algorithm for Spiking Neural Networks}},
  IEEE Transactions on Neural Networks and Learning Systems 28~(4) (2017)
  900--910.
\newblock \href {http://arxiv.org/abs/1512.01314} {\path{arXiv:1512.01314}},
  \href {https://doi.org/10.1109/TNNLS.2016.2582517}
  {\path{doi:10.1109/TNNLS.2016.2582517}}.
\newline\urlprefix\url{http://ieeexplore.ieee.org/document/7508492/}

\bibitem{Gerstner2002a}
W.~Gerstner, W.~M. Kistler,
  \href{https://www.cambridge.org/core/product/identifier/9780511815706/type/book}{{Spiking
  Neuron Models}}, Cambridge University Press, 2002.
\newblock \href {https://doi.org/10.1017/CBO9780511815706}
  {\path{doi:10.1017/CBO9780511815706}}.
\newline\urlprefix\url{https://www.cambridge.org/core/product/identifier/9780511815706/type/book}

\bibitem{Izhikevich2004}
E.~Izhikevich, \href{http://ieeexplore.ieee.org/document/1333071/}{{Which Model
  to Use for Cortical Spiking Neurons?}}, IEEE Transactions on Neural Networks
  15~(5) (2004) 1063--1070.
\newblock \href {https://doi.org/10.1109/TNN.2004.832719}
  {\path{doi:10.1109/TNN.2004.832719}}.
\newline\urlprefix\url{http://ieeexplore.ieee.org/document/1333071/}

\bibitem{Izhikevich2003a}
E.~M. Izhikevich, {Simple Model of Spiking Neurons, IEEE Transctions on Neural
  Networks}, IEEE Trans. Neural Networks 14~(6) (2003) 1569--1572.
\newblock \href {https://doi.org/10.1109/TNN.2003.820440}
  {\path{doi:10.1109/TNN.2003.820440}}.

\bibitem{Taherkhani2020}
A.~Taherkhani, A.~Belatreche, Y.~Li, G.~Cosma, L.~P. Maguire, T.~McGinnity,
  \href{https://doi.org/10.1016/j.neunet.2019.09.036
  https://linkinghub.elsevier.com/retrieve/pii/S0893608019303181
  http://www.ncbi.nlm.nih.gov/pubmed/31726331}{{A review of learning in
  biologically plausible spiking neural networks}}, Neural Networks 122 (2020)
  253--272.
\newblock \href {https://doi.org/10.1016/j.neunet.2019.09.036}
  {\path{doi:10.1016/j.neunet.2019.09.036}}.
\newline\urlprefix\url{https://doi.org/10.1016/j.neunet.2019.09.036
  https://linkinghub.elsevier.com/retrieve/pii/S0893608019303181
  http://www.ncbi.nlm.nih.gov/pubmed/31726331}

\bibitem{Chen2013}
J.~Y. Chen, P.~Lonjers, C.~Lee, M.~Chistiakova, M.~Volgushev, M.~Bazhenov,
  \href{https://www.jneurosci.org/lookup/doi/10.1523/JNEUROSCI.5088-12.2013}{{Heterosynaptic
  plasticity prevents runaway synaptic dynamics}}, Journal of Neuroscience
  33~(40) (2013) 15915--15929.
\newblock \href {https://doi.org/10.1523/JNEUROSCI.5088-12.2013}
  {\path{doi:10.1523/JNEUROSCI.5088-12.2013}}.
\newline\urlprefix\url{https://www.jneurosci.org/lookup/doi/10.1523/JNEUROSCI.5088-12.2013}

\bibitem{Shannon1948}
C.~E. Shannon, \href{https://ieeexplore.ieee.org/document/6773067}{{A
  Mathematical Theory of Communication}}, Bell System Technical Journal 27~(4)
  (1948) 623--656.
\newblock \href {https://doi.org/10.1002/j.1538-7305.1948.tb00917.x}
  {\path{doi:10.1002/j.1538-7305.1948.tb00917.x}}.
\newline\urlprefix\url{https://ieeexplore.ieee.org/document/6773067}

\bibitem{Yamaguchi2015}
Y.~Yamaguchi, M.~Miura, \href{http://dx.doi.org/10.1016/j.devcel.2015.01.019
  https://linkinghub.elsevier.com/retrieve/pii/S1534580715000623}{{Programmed
  Cell Death in Neurodevelopment}}, Developmental Cell 32~(4) (2015) 478--490.
\newblock \href {https://doi.org/10.1016/j.devcel.2015.01.019}
  {\path{doi:10.1016/j.devcel.2015.01.019}}.
\newline\urlprefix\url{http://dx.doi.org/10.1016/j.devcel.2015.01.019
  https://linkinghub.elsevier.com/retrieve/pii/S1534580715000623}

\bibitem{Thorpe1998}
S.~Thorpe, J.~Gautrais,
  \href{http://link.springer.com/10.1007/978-1-4615-4831-7{\_}19}{{Rank Order
  Coding}}, in: Computational Neuroscience, Springer US, Boston, MA, 1998, pp.
  113--118.
\newblock \href {https://doi.org/10.1007/978-1-4615-4831-7_19}
  {\path{doi:10.1007/978-1-4615-4831-7_19}}.
\newline\urlprefix\url{http://link.springer.com/10.1007/978-1-4615-4831-7{\_}19}

\bibitem{Delbruck2007}
T.~Delbruck, P.~Lichtsteiner,
  \href{http://ieeexplore.ieee.org/document/4252767/}{{Fast sensory motor
  control based on event-based hybrid neuromorphic-procedural system}}, in:
  2007 IEEE International Symposium on Circuits and Systems, no. 80 cm, IEEE,
  2007, pp. 845--848.
\newblock \href {https://doi.org/10.1109/ISCAS.2007.378038}
  {\path{doi:10.1109/ISCAS.2007.378038}}.
\newline\urlprefix\url{http://ieeexplore.ieee.org/document/4252767/}

\bibitem{Taylor2014}
D.~Taylor, N.~Scott, N.~Kasabov, E.~Capecci, E.~Tu, N.~Saywell, Y.~Chen, J.~Hu,
  Z.-G. Hou, \href{http://ieeexplore.ieee.org/document/6889936/}{{Feasibility
  of NeuCube SNN architecture for detecting motor execution and motor intention
  for use in BCIapplications}}, in: 2014 International Joint Conference on
  Neural Networks (IJCNN), IEEE, 2014, pp. 3221--3225.
\newblock \href {https://doi.org/10.1109/IJCNN.2014.6889936}
  {\path{doi:10.1109/IJCNN.2014.6889936}}.
\newline\urlprefix\url{http://ieeexplore.ieee.org/document/6889936/}

\bibitem{Koelstra2012}
S.~Koelstra, C.~Muhl, M.~Soleymani, {Jong-Seok Lee}, A.~Yazdani, T.~Ebrahimi,
  T.~Pun, A.~Nijholt, I.~Patras,
  \href{http://ieeexplore.ieee.org/document/5871728/}{{DEAP: A Database for
  Emotion Analysis ;Using Physiological Signals}}, IEEE Transactions on
  Affective Computing 3~(1) (2012) 18--31.
\newblock \href {https://doi.org/10.1109/T-AFFC.2011.15}
  {\path{doi:10.1109/T-AFFC.2011.15}}.
\newline\urlprefix\url{http://ieeexplore.ieee.org/document/5871728/}

\bibitem{Golmohammadi2019}
M.~Golmohammadi, A.~H. {Harati Nejad Torbati}, S.~{Lopez de Diego}, I.~Obeid,
  J.~Picone,
  \href{https://www.frontiersin.org/article/10.3389/fnhum.2019.00076/full}{{Automatic
  analysis of EEGs using big data and hybrid deep learning architectures}},
  Frontiers in Human Neuroscience 13 (2019) 76.
\newblock \href {http://arxiv.org/abs/1712.09771} {\path{arXiv:1712.09771}},
  \href {https://doi.org/10.3389/fnhum.2019.00076}
  {\path{doi:10.3389/fnhum.2019.00076}}.
\newline\urlprefix\url{https://www.frontiersin.org/article/10.3389/fnhum.2019.00076/full}

\bibitem{Watt2010}
A.~J. Watt, N.~S. Desai, \href{http://www.ncbi.nlm.nih.gov/pubmed/21423491
  http://www.pubmedcentral.nih.gov/articlerender.fcgi?artid=PMC3059670}{{Homeostatic
  Plasticity and STDP: Keeping a Neuron's Cool in a Fluctuating World.}},
  Frontiers in synaptic neuroscience 2~(JUN) (2010) 5.
\newblock \href {https://doi.org/10.3389/fnsyn.2010.00005}
  {\path{doi:10.3389/fnsyn.2010.00005}}.
\newline\urlprefix\url{http://www.ncbi.nlm.nih.gov/pubmed/21423491
  http://www.pubmedcentral.nih.gov/articlerender.fcgi?artid=PMC3059670}

\bibitem{Bell1995}
A.~J. Bell, T.~J. Sejnowski,
  \href{https://direct.mit.edu/neco/article/7/6/1129-1159/5909}{{An
  information-maximization approach to blind separation and blind
  deconvolution.}}, Neural computation 7~(6) (1995) 1129--1159.
\newblock \href {https://doi.org/10.1162/neco.1995.7.6.1129}
  {\path{doi:10.1162/neco.1995.7.6.1129}}.
\newline\urlprefix\url{https://direct.mit.edu/neco/article/7/6/1129-1159/5909}

\bibitem{Mellor2020}
J.~Mellor, J.~Turner, A.~Storkey, E.~J. Crowley,
  \href{http://arxiv.org/abs/2006.04647}{{Neural Architecture Search without
  Training}}, arXiv (jun 2020).
\newblock \href {http://arxiv.org/abs/2006.04647} {\path{arXiv:2006.04647}}.
\newline\urlprefix\url{http://arxiv.org/abs/2006.04647}

\bibitem{Bertschinger2004}
N.~Bertschinger, T.~Natschl{\"{a}}ger, {Real-time computation at the edge of
  chaos in recurrent neural networks}, Neural Computation 16~(7) (2004)
  1413--1436.
\newblock \href {https://doi.org/10.1162/089976604323057443}
  {\path{doi:10.1162/089976604323057443}}.

\bibitem{Li2017}
X.~Li, Q.~Chen, F.~Xue,
  \href{https://royalsocietypublishing.org/doi/10.1098/rsta.2016.0286}{{Biological
  modelling of a computational spiking neural network with neuronal
  avalanches}}, Philosophical Transactions of the Royal Society A:
  Mathematical, Physical and Engineering Sciences 375~(2096) (2017) 20160286.
\newblock \href {https://doi.org/10.1098/rsta.2016.0286}
  {\path{doi:10.1098/rsta.2016.0286}}.
\newline\urlprefix\url{https://royalsocietypublishing.org/doi/10.1098/rsta.2016.0286}

\bibitem{Willmore2001}
B.~Willmore, D.~Tolhurst,
  \href{https://www.tandfonline.com/doi/full/10.1080/net.12.3.255.270}{{Characterizing
  the sparseness of neural codes}}, Network: Computation in Neural Systems
  12~(3) (2001) 255--270.
\newblock \href {https://doi.org/10.1080/net.12.3.255.270}
  {\path{doi:10.1080/net.12.3.255.270}}.
\newline\urlprefix\url{https://www.tandfonline.com/doi/full/10.1080/net.12.3.255.270}

\bibitem{Abraham2019}
W.~C. Abraham, O.~D. Jones, D.~L. Glanzman,
  \href{http://dx.doi.org/10.1038/s41539-019-0048-y
  http://www.nature.com/articles/s41539-019-0048-y}{{Is plasticity of synapses
  the mechanism of long-term memory storage?}}, npj Science of Learning 4~(1)
  (2019) 9.
\newblock \href {https://doi.org/10.1038/s41539-019-0048-y}
  {\path{doi:10.1038/s41539-019-0048-y}}.
\newline\urlprefix\url{http://dx.doi.org/10.1038/s41539-019-0048-y
  http://www.nature.com/articles/s41539-019-0048-y}

\bibitem{Tavanaei2019}
A.~Tavanaei, M.~Ghodrati, S.~R. Kheradpisheh, T.~Masquelier, A.~Maida, {Deep
  learning in spiking neural networks}, Neural Networks 111~(January 2019)
  (2019) 47--63.
\newblock \href {http://arxiv.org/abs/2546440} {\path{arXiv:2546440}}, \href
  {https://doi.org/10.1016/j.neunet.2018.12.002}
  {\path{doi:10.1016/j.neunet.2018.12.002}}.

\bibitem{Kriener2014}
B.~Kriener, H.~Enger, T.~Tetzlaff, H.~E. Plesser, M.-O. Gewaltig, G.~T.
  Einevoll,
  \href{http://journal.frontiersin.org/article/10.3389/fncom.2014.00136/abstract}{{Dynamics
  of self-sustained asynchronous-irregular activity in random networks of
  spiking neurons with strong synapses}}, Frontiers in Computational
  Neuroscience 8~(October) (oct 2014).
\newblock \href {https://doi.org/10.3389/fncom.2014.00136}
  {\path{doi:10.3389/fncom.2014.00136}}.
\newline\urlprefix\url{http://journal.frontiersin.org/article/10.3389/fncom.2014.00136/abstract}

\bibitem{Mozzachiodi2010}
R.~Mozzachiodi, J.~H. Byrne,
  \href{https://linkinghub.elsevier.com/retrieve/pii/S0166223609001702}{{More
  than synaptic plasticity: role of nonsynaptic plasticity in learning and
  memory}}, Trends in Neurosciences 33~(1) (2010) 17--26.
\newblock \href {https://doi.org/10.1016/j.tins.2009.10.001}
  {\path{doi:10.1016/j.tins.2009.10.001}}.
\newline\urlprefix\url{https://linkinghub.elsevier.com/retrieve/pii/S0166223609001702}

\bibitem{Tian2021}
S.~Tian, L.~Qu, L.~Wang, K.~Hu, N.~Li, W.~Xu,
  \href{https://linkinghub.elsevier.com/retrieve/pii/S0925231221003325}{{A
  neural architecture search based framework for liquid state machine design}},
  Neurocomputing 443 (2021) 174--182.
\newblock \href {http://arxiv.org/abs/2004.07864} {\path{arXiv:2004.07864}},
  \href {https://doi.org/10.1016/j.neucom.2021.02.076}
  {\path{doi:10.1016/j.neucom.2021.02.076}}.
\newline\urlprefix\url{https://linkinghub.elsevier.com/retrieve/pii/S0925231221003325}

\bibitem{Beggs2003}
J.~M. Beggs, D.~Plenz,
  \href{https://www.jneurosci.org/lookup/doi/10.1523/JNEUROSCI.23-35-11167.2003}{{Neuronal
  Avalanches in Neocortical Circuits}}, Journal of Neuroscience 23~(35) (2003)
  11167--11177.
\newblock \href {https://doi.org/10.1523/jneurosci.23-35-11167.2003}
  {\path{doi:10.1523/jneurosci.23-35-11167.2003}}.
\newline\urlprefix\url{https://www.jneurosci.org/lookup/doi/10.1523/JNEUROSCI.23-35-11167.2003}

\bibitem{Schrauwen2008}
B.~Schrauwen, M.~Wardermann, D.~Verstraeten, J.~J. Steil, D.~Stroobandt,
  {Improving reservoirs using intrinsic plasticity}, Neurocomputing 71~(7-9)
  (2008) 1159--1171.
\newblock \href {https://doi.org/10.1016/j.neucom.2007.12.020}
  {\path{doi:10.1016/j.neucom.2007.12.020}}.

\bibitem{Lee2016}
J.~H. Lee, T.~Delbruck, M.~Pfeiffer,
  \href{http://journal.frontiersin.org/article/10.3389/fnins.2016.00508/full}{{Training
  Deep Spiking Neural Networks Using Backpropagation}}, Frontiers in
  Neuroscience 10 (nov 2016).
\newblock \href {http://arxiv.org/abs/1608.08782} {\path{arXiv:1608.08782}},
  \href {https://doi.org/10.3389/fnins.2016.00508}
  {\path{doi:10.3389/fnins.2016.00508}}.
\newline\urlprefix\url{http://journal.frontiersin.org/article/10.3389/fnins.2016.00508/full}

\bibitem{Markram2011a}
H.~Markram,
  \href{http://journal.frontiersin.org/article/10.3389/fnsyn.2011.00004/abstract}{{A
  history of spike-timing-dependent plasticity}}, Frontiers in Synaptic
  Neuroscience 3 (2011).
\newblock \href {https://doi.org/10.3389/fnsyn.2011.00004}
  {\path{doi:10.3389/fnsyn.2011.00004}}.
\newline\urlprefix\url{http://journal.frontiersin.org/article/10.3389/fnsyn.2011.00004/abstract}

\bibitem{Campanac2008}
E.~Campanac, G.~Daoudal, N.~Ankri, D.~Debanne,
  \href{https://www.jneurosci.org/lookup/doi/10.1523/JNEUROSCI.1411-08.2008}{{Downregulation
  of Dendritic Ih in CA1 Pyramidal Neurons after LTP}}, Journal of Neuroscience
  28~(34) (2008) 8635--8643.
\newblock \href {https://doi.org/10.1523/JNEUROSCI.1411-08.2008}
  {\path{doi:10.1523/JNEUROSCI.1411-08.2008}}.
\newline\urlprefix\url{https://www.jneurosci.org/lookup/doi/10.1523/JNEUROSCI.1411-08.2008}

\bibitem{VanWelie2004}
I.~{Van Welie}, J.~A. {Van Hoof}, W.~J. Wadman,
  \href{http://www.pnas.org/cgi/doi/10.1073/pnas.0307711101}{{Homeostatic
  scaling of neuronal excitability by synaptic modulation of somatic
  hyperpolarization-activated Ih channels}}, Proceedings of the National
  Academy of Sciences of the United States of America 101~(14) (2004)
  5123--5128.
\newblock \href {https://doi.org/10.1073/pnas.0307711101}
  {\path{doi:10.1073/pnas.0307711101}}.
\newline\urlprefix\url{http://www.pnas.org/cgi/doi/10.1073/pnas.0307711101}

\bibitem{Zhang2021}
W.~Zhang, P.~Li, \href{http://arxiv.org/abs/2108.01793}{{Composing Recurrent
  Spiking Neural Networks using Locally-Recurrent Motifs and Risk-Mitigating
  Architectural Optimization}} (aug 2021).
\newblock \href {http://arxiv.org/abs/2108.01793} {\path{arXiv:2108.01793}}.
\newline\urlprefix\url{http://arxiv.org/abs/2108.01793}

\bibitem{Golovko2003}
V.~Golovko, Y.~Savitsky, N.~Maniakov,
  \href{https://www.google.com/url?sa=t{\&}rct=j{\&}q={\&}esrc=s{\&}source=web{\&}cd=2{\&}ved=2ahUKEwibjqW2kOjeAhUJkywKHY-mAhwQFjABegQIBRAC{\&}url=https{\%}3A{\%}2F{\%}2Fpdfs.semanticscholar.org{\%}2F6f12{\%}2Fb4fa56f0018f6b92cdb47abed11c50b1606e.pdf{\&}usg=AOvVaw1NlT837CwLCGTgIEWCue4K}{{Neural
  Networks for Signal Processing in Measurement Analysis and Industrial
  Applications : the Case of Chaotic Signal Processing}}, Nato Science Series
  Sub Series III Computer and Systems Sciences 185 185 (2003) 119--144.
\newline\urlprefix\url{https://www.google.com/url?sa=t{\&}rct=j{\&}q={\&}esrc=s{\&}source=web{\&}cd=2{\&}ved=2ahUKEwibjqW2kOjeAhUJkywKHY-mAhwQFjABegQIBRAC{\&}url=https{\%}3A{\%}2F{\%}2Fpdfs.semanticscholar.org{\%}2F6f12{\%}2Fb4fa56f0018f6b92cdb47abed11c50b1606e.pdf{\&}usg=AOvVaw1NlT837CwLCGTgIEWCue4K}

\bibitem{Demin2018a}
V.~Demin, D.~Nekhaev,
  \href{https://www.frontiersin.org/article/10.3389/fninf.2018.00079/full}{{Recurrent
  spiking neural network learning based on a competitive maximization of
  neuronal activity}}, Frontiers in Neuroinformatics 12 (2018) 79.
\newblock \href {https://doi.org/10.3389/fninf.2018.00079}
  {\path{doi:10.3389/fninf.2018.00079}}.
\newline\urlprefix\url{https://www.frontiersin.org/article/10.3389/fninf.2018.00079/full}

\bibitem{Chen2018}
X.~Chen, W.~Kang, D.~Zhu, X.~Zhang, N.~Lei, Y.~Zhang, Y.~Zhou, W.~Zhao,
  \href{http://xlink.rsc.org/?DOI=C7NR09722K}{{A compact skyrmionic
  leaky–integrate–fire spiking neuron device}}, Nanoscale 10~(13) (2018)
  6139--6146.
\newblock \href {https://doi.org/10.1039/C7NR09722K}
  {\path{doi:10.1039/C7NR09722K}}.
\newline\urlprefix\url{http://xlink.rsc.org/?DOI=C7NR09722K}

\bibitem{Friedman2012}
N.~Friedman, S.~Ito, B.~A. Brinkman, M.~Shimono, R.~E. Deville, K.~A. Dahmen,
  J.~M. Beggs, T.~C. Butler,
  \href{https://link.aps.org/doi/10.1103/PhysRevLett.108.208102}{{Universal
  critical dynamics in high resolution neuronal avalanche data}}, Physical
  Review Letters 108~(20) (2012) 208102.
\newblock \href {https://doi.org/10.1103/PhysRevLett.108.208102}
  {\path{doi:10.1103/PhysRevLett.108.208102}}.
\newline\urlprefix\url{https://link.aps.org/doi/10.1103/PhysRevLett.108.208102}

\end{thebibliography}

\newpage

\appendix

\section{Spiking raster plots under different learning parameter settings}
\label{overActive}
\begin{figure}[!h]
	\begin{center}
		\subfloat[\label{12a}]{%
			\includegraphics[width=0.5\linewidth]{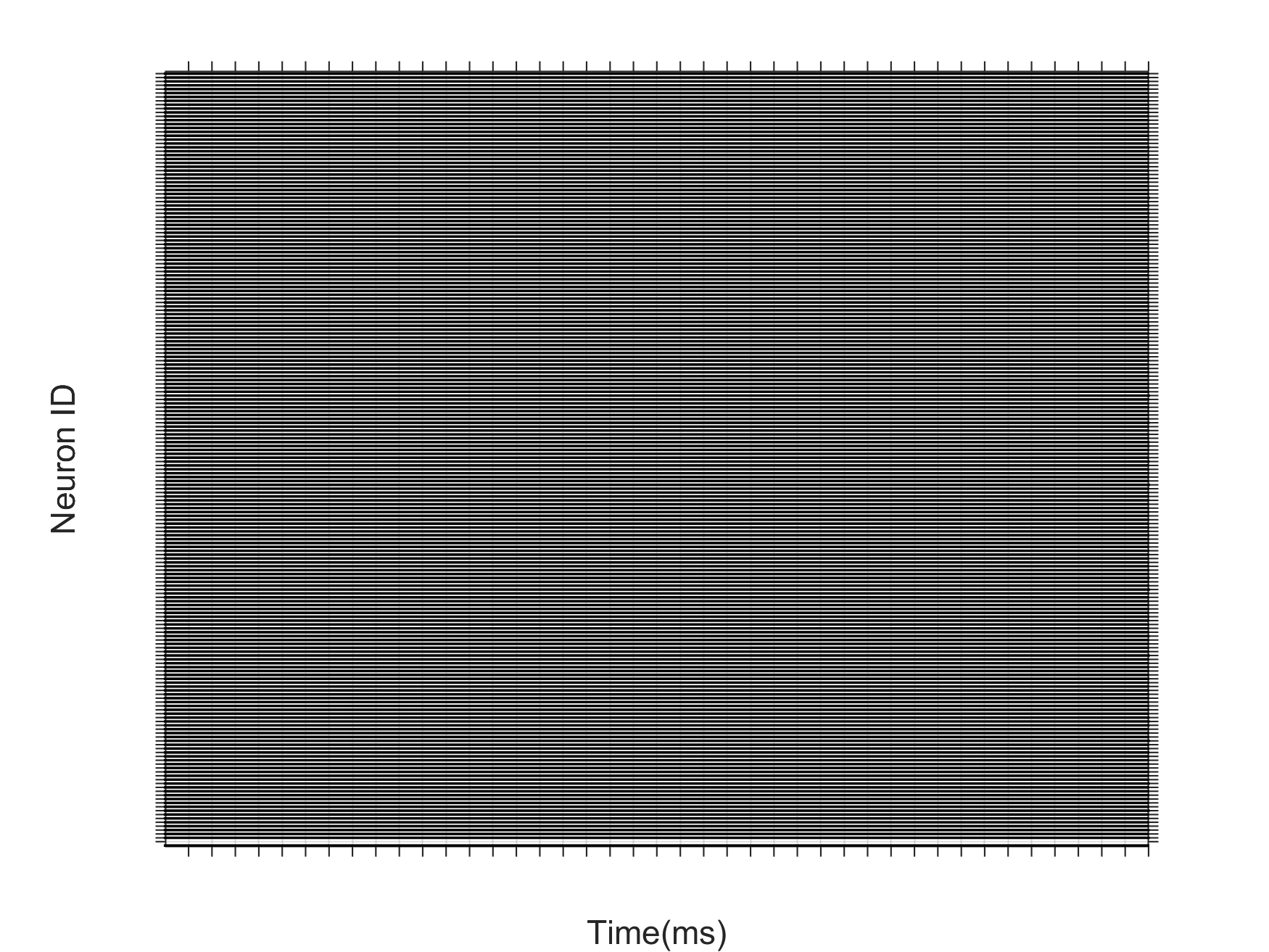}%
				}\hfill
		\subfloat[\label{12b}]{%
			\includegraphics[width=0.5\linewidth]{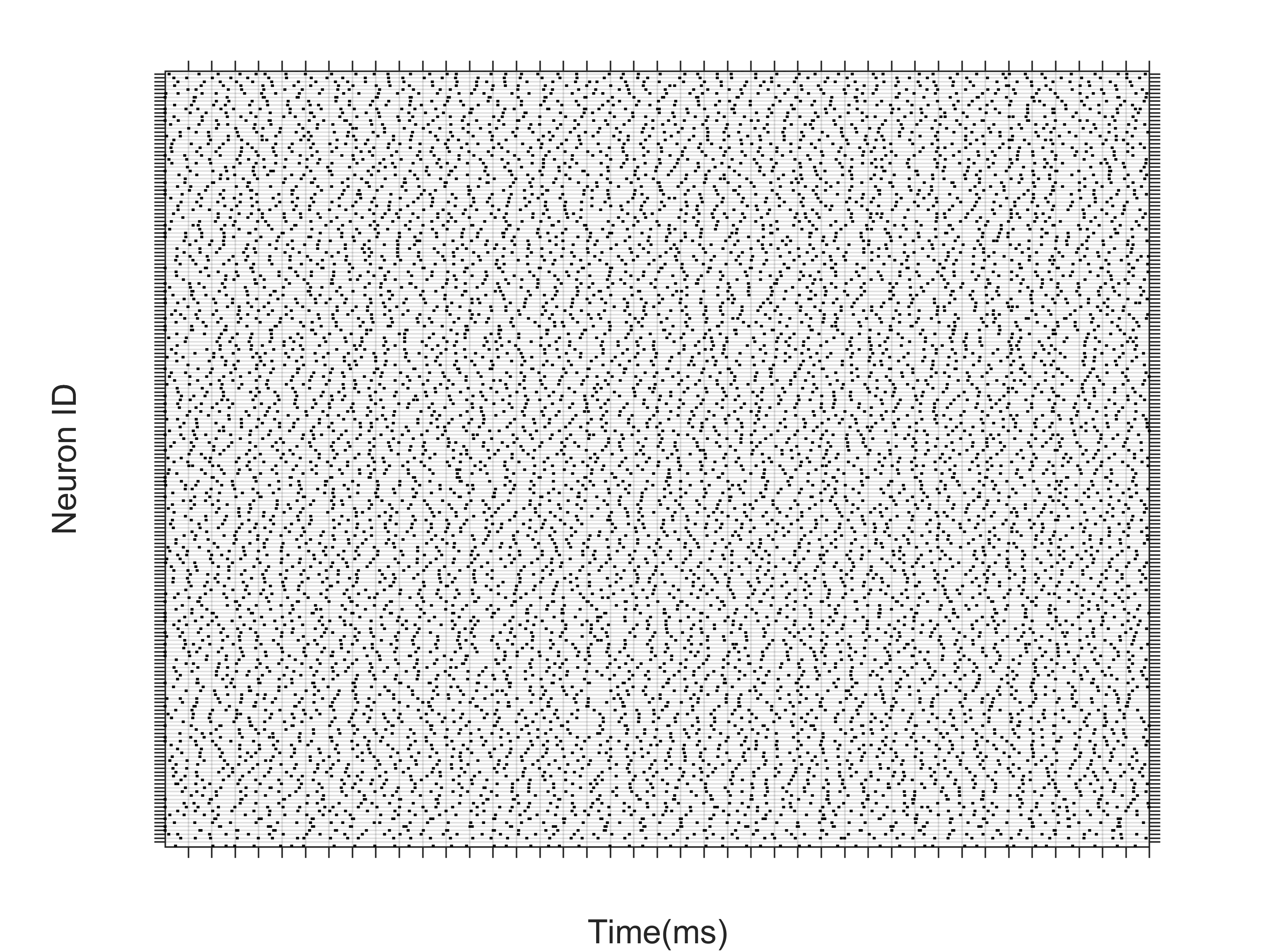}%
				}
		\captionof{figure}{Spiking raster plots under different $\theta_{pos}$ and $\theta_{neg}$ conditions a) $\theta_{pos} < \theta_{neg}$  b) $\theta_{pos} = \theta_{neg}$ }
		\label{DifIpTune}
	\end{center}
\end{figure}
\FloatBarrier

\section{AER algorithm}
\label{AER}
\begin{algorithm}
	
	\SetKwInOut{Input}{Input}
	\SetKwInOut{Output}{Output}
	
	\underline{function AER} $(x_{(t)},f)$\;
	\Input{EEG channel data $x_{(t)}$ and factor $f$}
	\Output{Spike train $s_{(t)}$}
	\For{$t=1:length(x_{(t)})$}
	{
		$tempdiff_{(t)} = x_{(t+1)} - x_{(t)}$
		
		return $tempdiff_{(t)}$;
	}
	\text{$tempdiff_{(end)}=tempdiff_{(end-1)}$}
	
	\text{$threshold=mean(tempdiff_{(t)})+(f*std(tempdiff_{(t)}))$}
	
	\For{$t=1:length(x_{(t)})$}
	{
		\eIf{$tempdiff_{(t)}>threshold$}{$s_{(t)}=1$}{\If{$tempdiff_{(t)}<threshold$}{$s_{(t)}=-1$}}
		
		return $s_{(t)}$\;
	}
	\caption{AER encoding algorithm for spike conversion}
\end{algorithm} 

\newpage
\section{Hyper-parameter settings}
\label{Hyperparameter}
\begin{table}[!h]
	\begin{threeparttable}

		\caption{Hyper-parameter values used for experimentation}
		
		\centering
		\begin{tabular}{c c c
				%				*{2}{S[table-format=1.2]}
				%				S[table-format=2.2] 
				%				*{3}{S[table-format=1.2]}
				%				S[table-format=2.0]
			}
			\toprule
			\multicolumn{1}{c}{Processing step}
			& \multicolumn{1}{c}{Hyper-parameter}
			& \multicolumn{1}{l}{Value} \\
			
			%			\multicolumn{1}{c}{Processing step}
			%			& \multicolumn{1}{c}{Hyper-parameter}
			%			& \multicolumn{1}{c}{Value} \\
			
			\midrule
			
			AER Encoder	&$factor$&	0.5\\
			\hline
			LIF		&$v_{init}$&	0.05\\
			&$v_{rest}$&	0\\
			&$t_{refractory}$&	5\\
			&$R$&	1\\
			&$C$&	10\\ 
			&$\eta_{init}$\tnote{a}&	200\\
			&$\eta_{init}$\tnote{b}&	300\\ 		   
			\hline
			STDP	&$A_{+}$&	0.001\\
			&$A_{-}$&	0.001\\
			&$\tau_{pos}$&	10\\
			&$\tau_{neg}$&	10\\
			&$w_{max}$&	+0.1\\
			&$w_{min}$&	-0.1\\
			\hline
			IP	   	&$\theta_{pos}$\tnote{a}&	\num{1E-3}\\
			&$\theta_{neg}$\tnote{a}&	\num{1E-5}\\
			&$\theta_{pos}$\tnote{b}&	\num{7E-5}\\
			&$\theta_{neg}$\tnote{b}&	\num{7E-8}\\
			\hline
			Classifier	   	&$\alpha$&	1\\
			&$mod$&	0.8\\
			&$drift$&	0.001\\			
			\bottomrule
			
		\end{tabular}
		\begin{tablenotes}
			\item[a]{For Wrist Flexion data set modeling}
			\item[b]{For DEAP data modeling}
			
		\end{tablenotes}
	\end{threeparttable}
\end{table}




%%\bibitem{}

%\end{thebibliography}
\end{document}